\documentclass[sigconf, nonacm]{acmart}

\pdfoutput=1

\usepackage[a-2b]{pdfx}
\usepackage{adjustbox}

\usepackage{caption}
\usepackage{multirow}
\usepackage{multicol}
\usepackage{enumitem}
\usepackage{rotating}
\usepackage{siunitx}
\usepackage{algorithm}
\usepackage{colortbl}
\usepackage{xcolor}

\usepackage{listings}
\usepackage{xcolor}

\definecolor{codegreen}{rgb}{0,0.6,0}
\definecolor{codegray}{rgb}{0.5,0.5,0.5}
\definecolor{codepurple}{rgb}{0.58,0,0.82}
\definecolor{backcolour}{rgb}{0.95,0.95,0.92}

\lstdefinestyle{mystyle}{
    backgroundcolor=\color{backcolour},   
    commentstyle=\color{codegreen},
    keywordstyle=\color{magenta},
    numberstyle=\tiny\color{codegray},
    stringstyle=\color{codepurple},
    basicstyle=\ttfamily\footnotesize,
    breakatwhitespace=false,         
    breaklines=true,                 
    captionpos=b,                    
    keepspaces=true,                 
    numbers=left,                    
    numbersep=5pt,                  
    showspaces=false,                
    showstringspaces=false,
    showtabs=false,                  
    tabsize=2
}

\usepackage{subfig}
\usepackage{extra_commands}

\begin{document}
\title{Locally-Adaptive Quantization for Streaming Vector Search}

\author{Cecilia Aguerrebere}
\affiliation{%
  \institution{Intel Labs}
}
\email{cecilia.aguerrebere@intel.com}

\author{Mark Hildebrand}
\affiliation{%
  \institution{Intel Labs}
}
\email{mark.hildebrand@intel.com}

\author{Ishwar Singh Bhati}
\affiliation{%
  \institution{Intel Labs}
}
\email{ishwar.s.bhati@intel.com}

\author{Theodore Willke}
\affiliation{%
  \institution{Intel Labs}
}
\email{ted.willke@intel.com}

\author{Mariano Tepper}
\affiliation{%
  \institution{Intel Labs}
}
\email{mariano.tepper@intel.com}

\begin{abstract}
Retrieving the most similar vector embeddings to a given query among a massive collection of vectors has long been a key component of countless real-world applications. The recently introduced Retrieval-Augmented Generation is one of the most prominent examples. For many of these applications, the database evolves over time by inserting new data and removing outdated data. In these cases, the retrieval problem is known as streaming similarity search. While Locally-Adaptive Vector Quantization (LVQ), a highly efficient vector compression method, yields state-of-the-art search performance for non-evolving databases, its usefulness in the streaming setting has not been yet established. In this work, we study LVQ in streaming similarity search. In support of our evaluation, we introduce two improvements of LVQ: Turbo LVQ and multi-means LVQ that boost its search performance by up to 28\% and 27\%, respectively. Our studies show that LVQ and its new variants enable blazing fast vector search, outperforming its closest competitor by up to 9.4x for identically distributed data and by up to 8.8x under the challenging scenario of data distribution shifts (i.e., where the statistical distribution of the data changes over time).  We release our contributions as part of Scalable Vector Search, an open-source library for high-performance similarity search.
\end{abstract}

\maketitle

\section{Introduction}

Similarity search, the process of retrieving from a massive collection of vectors those that are most similar to a given query vector, is a key component of countless classical real-world applications (e.g., recommender systems or ad matching). In recent years, the use of similarity search has grown exponentially with the rise of deep learning models that can translate semantic affinities into spatial similarities~\cite{devlin_bert_2019,radford_learning_2021,brown_language_2020} and consequently enable semantic search. A prominent example is Retrieval-Augmented Generation (RAG)~\cite{lewis_retrieval-augmented_2020} that extends the outstanding capabilities of Generative Artificial Intelligence (AI)~\cite{cai_recent_2022,liu_retrieval-augmented_2021,jiang_active_2023} with more factually accurate, up-to-date, and verifiable results. Although in virtually all deployments of these applications the database changes over time, the scientific literature has devoted surprisingly little attention to \emph{streaming similarity search}, where the database is built dynamically by adding and removing vectors.

Streaming similarity search comes with its own unique challenges. Whereas keeping data-agnostic indices (e.g., LSH~\cite{datar_locality-sensitive_2004,gionis_similarity_1999}) up to date is trivial, their accuracy and speed falter. Data-driven indices, among which graph-based approaches dominate, stand out by offering fast and highly-accurate search for billions of high-dimensional vectors~\cite{aguerrebere_similarity_2023,shimomura_survey_2021,li_approximate_2020}. However, incrementally updating the internal data structures while keeping the vector search fast and accurate is still an open research problem. Moreover, and depending on the application, the vectors in the data stream can be independent and identically distributed (IID) or may undergo a data distribution shift over time, making the problem even more challenging. Think, for example, of a retail company changing its product catalogue over time or of seasonal/cultural trends causing content drifts in social media.

Scalable Vector Search (\svsfullp{}),\footnote{\url{https://github.com/IntelLabs/ScalableVectorSearch}} is a recently introduced open-source library for data-driven similarity search that is state-of-the-art in the \emph{static} setting (where the database is fixed and never updated), outperforming its competitors by up to 5.8x~\cite{aguerrebere_similarity_2023}. One of the main underpinnings of \svsfullp{}' performance is Locally-adaptive Vector Quantization (LVQ)~\cite{aguerrebere_similarity_2023}, a highly efficient vector compression method that accelerates vector similarity computations, reduces the memory bandwidth consumption, and decreases the memory footprint, all with no significant impact in search accuracy. 
As LVQ relies on \emph{global data statistics} to compress each vector individually and these statistics can change over time for streaming data, the quality of the LVQ representation and, consequently, of the similarity searches can potentially be affected in unforeseen ways. This work is devoted to determining how to achieve the benefits of LVQ in the streaming setting.

Specifically, for a database of vectors $\{ \vect{x}_i \}_{i=1}^{n} $, LVQ uses the sample mean $\vect{\mu} = \frac{1}{n} \sum_{i=1}^{n} \vect{x}_i$ to homogenize the distributions across vector dimensions. Although effective in the static case, the initial estimate can be arbitrarily inaccurate (1) when $\vect{\mu}$ is computed from a small initial set of vectors, and/or (2) under distribution shifts as the data stream progresses. We conduct a thorough evaluation and analysis of these scenarios, aided with an experimental framework that relates LVQ compression errors to search accuracy.

Guided by the results from our analysis, we present two improvements.
\textbf{Turbo LVQ} boosts distance calculation performance  by modifying the underlying layout of the vector data to streamline its use with SIMD instructions.
We also augment LVQ with multiple local means, rendering the representation \emph{local} instead of global. Compared to LVQ, \textbf{Multi-Means LVQ} reduces the compression error, potentially improving the search accuracy.

In summary, this work presents the following contributions:
\begin{itemize}[topsep=0.ex,leftmargin=5ex]
    \item We empirically show that LVQ is robust to variations in the data distribution and yields vast performance gains in streaming similarity search over the state of the art, irrespective of the presence of data distribution shifts. \svs{} outperforms its closest competitor by up to 9.4x in the IID case and by up to 8.8x under data distribution shifts.
    \item We present two LVQ variants. Compared to vanilla LVQ, Turbo LVQ boosts search performance consistently by $28\%$ while Multi-Means LVQ, depending on the dataset, obtains speedups of up to 27\%.
    \item To ensure reproducibility, we incorporate the streaming techniques introduced in this work to Scalable Vector Search, an open-source library for high-performance similarity search.
    \item We introduce the first open-source dataset to evaluate streaming search techniques under data distribution shifts \footnote{Available at \url{https://github.com/IntelLabs/VectorSearchDatasets}}.
\end{itemize}

The manuscript is organized as follows. \cref{sec:problem_statement} introduces core concepts and states the main research questions. In \cref{sec:deep_dive_LVQ} we answer these questions by conducting an in-depth analysis of LVQ and also introduce two novel variants that boost its search performance. An exhaustive experimental evaluation supporting our contributions is presented in \cref{sec:experimental_evaluation}. We finish describing the related work in \cref{sec:related_work} and presenting a summary of our work in \cref{sec:conclusions}.

\section{Background and Problem Statement}
\label{sec:problem_statement}

\subsection{Streaming similarity search}
\label{ssec:streaming_simsearch}

In the streaming setting, we have an initial database $\X_0 = \{ \vect{x}_i \in \Real^d \}_{i=1}^{n_0}$, containing $n_0$ vectors in $d$ dimensions. Then, at each time $t > 1$, a vector $\vect{x}_i$ is either added or removed, i.e., $\X_t = \X_{t - 1} \cup \{ \vect{x}_i \}$ or $\X_t = \X_{t - 1} \setminus \{ \vect{x}_i \}$. We assume with no loss of generality that each vector has a universally unique ID.

Given $\X_t$, a symmetric similarity function $\simfun : \Real^d \times \Real^d \to \Real$ where a higher value indicates a higher degree of similarity, and a query $\q \in \Real^d$, the similarity search (or nearest neighbor) problem consists in finding the $k$ vectors in $\X_t$ with maximum similarity to $\q$.
In most practical applications, some accuracy is traded for performance to avoid a linear scan of $\X_t$, by relaxing the definition to allow for a certain degree of error, i.e., a few of the retrieved elements (the approximate nearest neighbors, ANN) may not belong to the ground-truth top $k$ neighbors.
Here, search accuracy is commonly measured by $k$-recall$@k$, defined by $| S \cap G_t | / k$, where $S$ are the IDs of the $k$ retrieved neighbors and $G_t$ is the ground-truth at time $t$. Unless otherwise specified, we use $k=10$ in all experiments and 0.9 as the default accuracy value. Search performance is measured in queries per second (QPS).

An ANN index enables ANN searches in $\X_t$. Given the lengthy index construction times of modern techniques, creating an ANN index from scratch at each time $t$ is prohibitive. Thus, the goal in the streaming scenario is to build and maintain for $t=1, \cdots$ a \emph{dynamic} index that is both accurate and fast. Streaming similarity search is a fundamental problem in real-world scenarios that require updating the database over time. For example, in recommendation systems or web search where similarity search is extensively used, new content is created and old content is discarded with high frequency.

\textbf{Data distribution shifts.}
In some applications, the distribution of the vectors may change over time. Consider, for example, a retail company incorporating new categories of products (see \cref{{fig:open-images-dataset}}), a retrieval-based multilingual large language model supporting new languages, or a retrieval-enhanced coding co-pilot supporting new programming languages. In these scenarios, vectors with potentially different embedding distributions will need to be indexed and searched for over time. We are interested here in \emph{natural} data distribution shifts where the deep learning model generating the embeddings is fixed, and the shift comes from the diversity of the model inputs~\cite{miller_effect_2020,taori_measuring_2020}. The study of scenarios where the model changes is out of the scope of this work.  

\begin{figure}
  \centering
  \begin{minipage}{0.58\columnwidth}
    \centering
    \includegraphics[width=\textwidth]{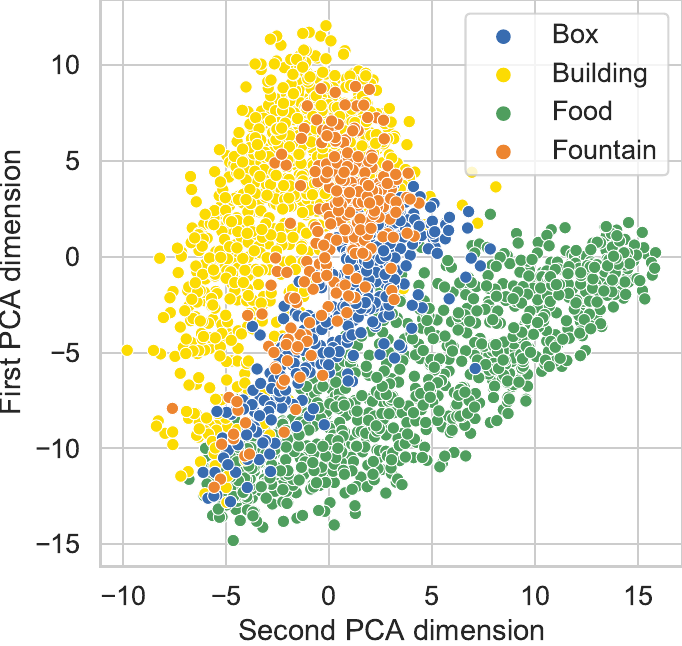}    
  \end{minipage}%
  \hfill%
  \begin{minipage}{0.4\columnwidth}
    \centering
    \includegraphics[width=\textwidth]{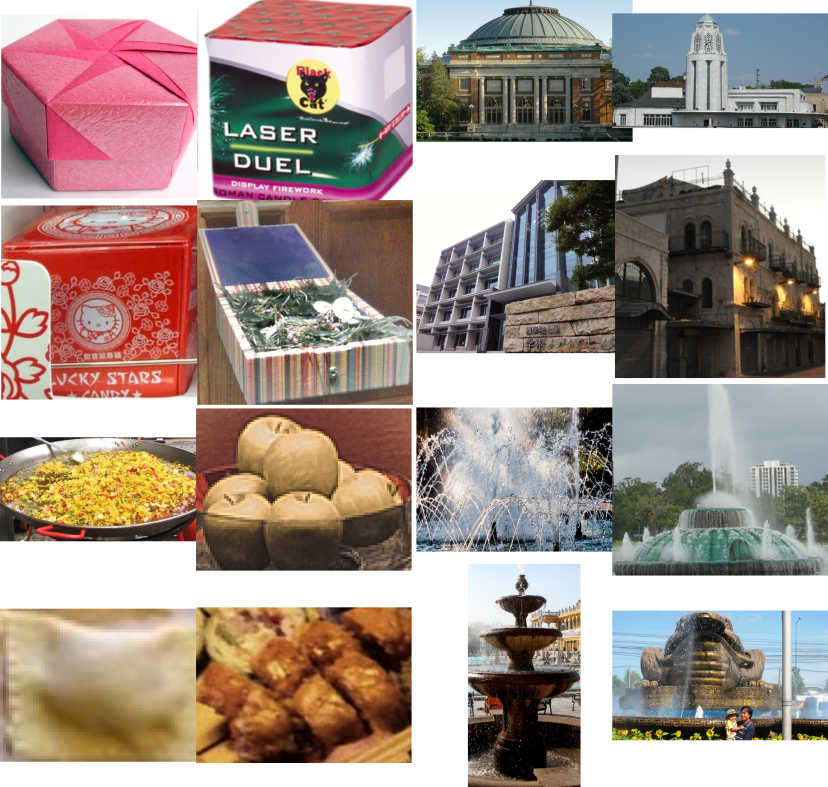}    
  \end{minipage}

  \caption{The newly introduced open-images-512-13M dataset (\cref{ssec:datasets_and_procols}) contains vectors organized in semantical classes that have different distributions, as evidenced by the 2D Principal Component Analysis projections of vectors sampled from four classes (left). For example, vectors from \texttt{fountain} are close to those from \texttt{building} as these images often display buildings.
  On the right, we show a few examples~\cite{kuznetsova_open_2020} from each of the four classes: \texttt{box} (top-left), \texttt{building} (top-right), \texttt{food} (bottom-left),  and \texttt{fountain} (bottom-right).%
  }
  \label{fig:open-images-dataset}
\end{figure}

\subsection{Graph-based streaming similarity search}
\label{ssec:graph_based_streaming_search}

Graph-based methods provide fast and highly accurate similarity search and constitute the state-of-the-art in both the \emph{static}~\cite{aguerrebere_similarity_2023} and dynamic~\cite{singh_freshdiskann_2021} cases. These indices build a proximity graph, connecting two nodes if they fulfill a defined neighborhood criterion with demonstrable properties~\cite{fu_fast_2019}, and use a greedy traversal to find the nearest neighbor~\cite{fu_fast_2019,malkov_efficient_2020,subramanya_diskann_2019}.

For search, the graph is traversed using a modified greedy best-first approach (details in \cref{app:graph_streaming_sim_search}) to retrieve the $k$ approximate nearest vectors to query $\q$ with respect to the similarity function $\simfun$. The search window size $W$ is a hyperparameter controlling the amount of backtracking allowed throughout the greedy traversal: increasing $W$ improves the accuracy of the retrieved neighbors by exploring more of the graph at the cost of an increased search time. 

Graph construction involves building a navigable graph for $\X_0$ and performing additions and deletions to update it over time. In this work, we use the FreshVamana~\cite{singh_freshdiskann_2021} dynamic index for its strong and stable performance, but our results apply to other graphs-based methods~\cite[e.g.,][]{malkov_efficient_2020}. A comprehensive description of the construction process is available in \cref{app:graph_streaming_sim_search}.

\subsection{Locally-adaptive Vector Quantization}
\label{ssec:introduce_lvq}
Locally-adaptive vector quantization (LVQ)~\cite{aguerrebere_similarity_2023} is a compression technique that uses per-vector scaling and scalar quantization to boost search performance by enabling blazingly fast similarity computations and a reduced effective bandwidth, while decreasing memory footprint and hardly impacting accuracy.

Let $\vect{\mu} = [\mu_1, \dots, \mu_d]$ be the sample mean, $\vect{\mu} = \frac{1}{n} \sum_{i=1}^{n} \vect{x}_i$, and $u, \ell : \Real^d \to \Real$ defined, for a vector $\vect{x} = [ x_1, \dots, x_d ]$, as
\begin{align}
    u(\vect{x}) &= \max_{j} x_j - \mu_j,
    &
    \ell(\vect{x}) &= \min_{j} x_j - \mu_j.
\label{eq:quant_bounds}
\end{align}
Let $Q_{B, \ell, u} : \Real \to \Real$ be the scalar quantization function,
\begin{equation}
    Q_{B, \ell, u}(x) = \Delta \left\lfloor \frac{x - \ell}{\Delta} + \frac{1}{2} \right\rfloor + \ell ,
    \quad\text{with}\quad
    \Delta = \tfrac{u - \ell}{2^B - 1}.
\label{eq:scalar_quant}
\end{equation}
In Locally-adaptive Vector Quantization (LVQ), the vector $\vect{x}$ is represented by a vector $Q(\vect{x})$ and, optionally, by another vector $Q_{\text{res}}(\vect{r})$, obtained by:
\begin{itemize}[topsep=0.ex,leftmargin=3ex]
    \item performing a first-level encoding of $\vect{x}$ into $Q(\vect{x})$ with $B_1$ bits using
    \begin{equation}
        Q(\vect{x}) = Q_{B_1, \ell(\vect{x}), u(\vect{x})}( \vect{x} - \vect{\mu}) ,
        \label{eq:quant_def}
    \end{equation}
    by applying $Q_{B, \ell, u}$ component-wise;
    \item optionally performing a second-level encoding of the residual vector $\vect{r} = \vect{x} - \vect{\mu} - Q(\vect{x})$ into $Q_{\text{res}}(\vect{r})$ with $B_2$ bits by applying $Q_{B_2, -\Delta/2, \Delta/2}$ component-wise (the components of $\vect{r}$ lie in $[-\Delta/2, \Delta/2]$).
\end{itemize}
Throughout this work, we denote the one-level and the two-level variants by LVQ-$B$ and LVQ-$B_1\mathsf{x}B_2$, respectively.

LVQ is designed in particular for graph-based similarity search, with its random memory access pattern~\cite{aguerrebere_similarity_2023}.
The first-level LVQ vectors are used during graph traversal, which improves the search performance by compressing the vectors into fewer bits and thus reducing the memory bandwidth effectively consumed. Any degradation in search accuracy caused by the scalar quantization errors can be regained by increasing the search window size \searchWin{} (\cref{ssec:graph_based_streaming_search}), at the cost of slowing down the search, and/or by using the second-level residuals to perform a final re-ranking. Finding the optimal configuration for each dataset boils down to finding small values for $B_1$ and $B_2$ such that $W$ does not need to be increased too much.

Following~\cite{aguerrebere_similarity_2023}, we build our graphs directly from first-level LVQ encodings, which does not affect the graph quality measured by search accuracy and performance. 

\subsubsection{The challenge of LVQ for streaming similarity search}
LVQ uses the sample mean $\vect{\mu}$ to homogenize the distributions across vector dimensions. Here, the implicit hypothesis is that the underlying generative model produces vectors with component-wise distributions that share the same span after subtracting the mean.
Subtracting the global mean works well in the \emph{static} similarity search case, where we have a large set of vectors, known beforehand. There, the sample mean becomes an accurate estimate by the law of large numbers and the component-wise distributions are similar. 

In the streaming case, however, the index may be initialized with a small number of vectors thus producing an inaccurate estimate of the mean. Moreover, in the case of data distribution shifts, the mean of the vectors may drift over time, decreasing the accuracy of the initial sample mean even further.
Updating $\vect{\mu}$ over time is a possible solution that would require entirely re-encoding the dataset each time, which may be prohibitive for large databases.

In these conditions, the component-wise distributions may start differing significantly from one another, violating the implicit LVQ model and consequently increasing its compression error. In this work, we analyze how sensitive LVQ is to vector de-meaning, and in particular how this impacts search accuracy and performance for streaming similarity search (see \cref{sec:analyzing_lvq} and \cref{sec:experimental_evaluation}).

\section{A deep dive on LVQ}
\label{sec:deep_dive_LVQ}
We now elaborate an in-depth analysis of the state-of-the-art LVQ, from its implementation to its performance at divergent quantization error regimes.
First, we present Turbo LVQ, a novel reformulation that boosts search performance by permuting the vector's memory layout for its use with SIMD instructions available in modern CPUs.
Second, we introduce multi-means LVQ (M-LVQ) that provides additional accuracy by using a localized statistical model in replacement of the global sample mean in LVQ. Finally, we provide a detailed study that covers the impact in search accuracy of the quantization granularity in LVQ and M-LVQ as well as the accuracy in the sample mean estimate.

\subsection{Turbo LVQ}
\label{ssec:turbo_implementation}
The original LVQ stores consecutive logical dimensions sequentially in memory. While convenient, this choice requires significant effort to unpack encoded dimensions into a more useful form. With Turbo LVQ, we recognize that consecutive logical dimensions need not be stored consecutively in memory and permute their order to facilitate faster decompression with SIMD instructions~\cite{afroozeh_fastlanes_2023}. 
In the following, we illustrate the ideas behind Turbo LVQ using $B_1=4$, the mechanism for $B_1=8$ being conceptually similar.

\begin{figure*}
    \centering
    \includegraphics[width=0.8\textwidth]{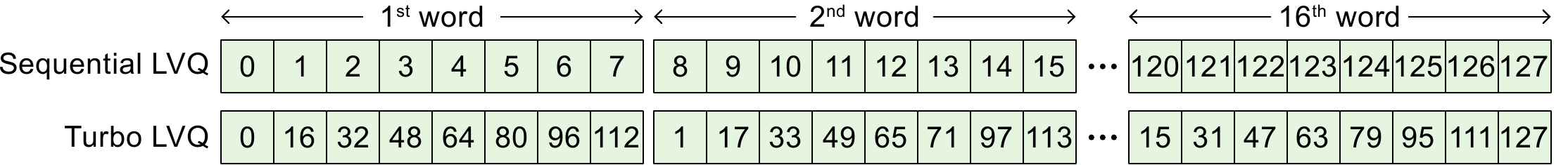}
    \caption{LVQ uses the original dimension ordering to store each vector. Turbo LVQ permutes the dimensions of the database vectors to make their decoding more efficient. Here, we show an example where the permutation is applied in groups of 128 dimensions, using 4 bits per dimension bundled in sixteen 32-bit words.}
    \label{fig:TurboLVQ_layout}
\end{figure*}

Let us begin by describing how the original version of LVQ works. LVQ stores consecutive logical dimensions sequentially in memory in 32-bit words as groups of eight 4-bit dimensions, see \cref{fig:TurboLVQ_layout}. Before doing any actual distance computations, each pair of words needs to be unpacked into a SIMD vector with 512 bits, containing sixteen 32-bit unsigned integers. Once we have the data in this unpacked format, we can undo the quantization and perform the partial similarity computation for these sixteen dimensions. In total, every unpacking of sixteen dimensions can be done using 7 assembly instructions (see \cref{app:turbo_implementation} for additional details).

Turbo LVQ uses a permuted memory layout storing groups of 128 dimensions, each encoded with 4 bits, into 64 bytes of memory. Dimension 0 is stored in the first 4 bits of the first register lane, dimension 1 is stored in the first 4 bits of the second register lane, continuing the pattern until dimension 16 which is stored in the second 4 bits of the first register lane, see \cref{fig:TurboLVQ_layout}. When decoding, the entire 64-bytes block is loaded into an AVX-512 register as 16 lanes of 32-bit integers. Then, the first sixteen dimensions (the lowest 4 bits of each word in \cref{fig:TurboLVQ_layout}) are extracted by simply applying a bitwise mask to each lane. For subsequent groups, a shift needs to be applied before recovering the lowest 4 bits. With this strategy, unpacking sixteen-dimensions requires only 2 assembly instructions: a load+mask for the first group and a shift+mask for each following group (see \cref{app:turbo_implementation} for additional details). All in all, Turbo LVQ, with its computational savings, pushes the graph-based search problem even further into its natural memory-limited regime~\cite{aguerrebere_similarity_2023}.

\subsection{Multi-Means LVQ}

LVQ characterizes the data distribution with its global mean. A natural idea for improving the accuracy of LVQ would be to model the data distribution more tightly. One simple way to achieve this improvement is to use a mixture model, where each vector is described with respect to a local mean.
We thus propose to augment LVQ with multiple means, which corresponds to a Gaussian mixture model with spherical components of equal variance.
In this setting, each vector is assigned to one of $M$ centers $\{ \vect{\mu}_m \}_{m=1}^{M}$, which are computed using k-means. We can then replace \cref{eq:quant_def} with $Q_{B_1, \ell(\vect{x}), u(\vect{x})} \left( \vect{x} - \vect{\mu}_{*} \right)$ where $\vect{\mu}_{*}$ is the closest center to $\vect{x}$, i.e., $\vect{\mu}_{*} = \argmin_m \| \vect{x} - \vect{\mu}_m \|_2^2$. Notice that the original LVQ is now a particular case where $M=1$.

\begin{definition}
\label{def:mlvq}
Let $\{ \vect{\mu}_m \in \Real^d \}_{m=1}^{M}$ be a collection of $M$ centers.
We define the multi-means LVQ compression ($M$-LVQ-$B_1\mathsf{x}B_2$) of vector $\vect{x}$, respectively with $B_1$ and $B_2$ bits for the first and second levels, as the pair of vectors $Q(\vect{x})$ and $Q_{\text{res}}(\vect{r})$ such that
\begin{itemize}[topsep=0.ex,leftmargin=3ex]
    \item $Q(\vect{x}) = Q_{B_1, \ell(\vect{x}), u(\vect{x})} \left( \vect{x} - \vect{\mu}_{*} \right)$ \enspace with \enspace $\displaystyle \vect{\mu}_{*} = \argmin_{m=1,\dots,M} \norm{\vect{x} - \vect{\mu}_m}{2}^2$,
    
    \item $Q_{\text{res}}(\vect{r}) = Q_{B_2, -\Delta/2, \Delta/2} (\vect{r})$ \enspace for \enspace $\vect{r} = \vect{x} - \vect{\mu}_{*} - Q(\vect{x})$,
    
\end{itemize}
where the scalar quantization function in \cref{eq:scalar_quant} is applied component-wise.
\end{definition}

The main benefit of M-LVQ in the static case is the reduction of the quantization error. As $M$ gets larger, $\| \vect{x} - \vect{\mu}_* \|_2^2$ goes to zero, making the quantization of $\vect{x} - \vect{\mu}_*$ more accurate. This, in turn, means that we can potentially reduce the search window size $W$ to speed-up the search. 

In the streaming case, M-LVQ is additionally motivated by an enhanced flexibility to update the encoding. As vectors are assigned to local centers, any additions will only trigger a re-encoding of the database vectors that are assigned to the same center as the new vector. Thus, we can update the centers as we process the stream, avoiding the need to entirely re-encode the database. This feature is also useful in the presence of large data distribution shifts, where the distribution drift is often local (e.g., when adding a new product class) and thus most centers can remain untouched.

Interestingly, we find that a small fraction of the $M$ centers are actually used during graph search for a query $\vect{q}$. This situation is analogous to the one encountered when using an inverted index for similarity search, where the data is clustered and the search is conducted by doing linear scans on a subset of the clusters that are closest to the query. As such, we can view graph search with M-LVQ as a \emph{soft inverted index}, where we avoid doing a linear scan of all the points assigned to each center $\vect{\mu}_m$, with the graph traversal performing a more fine-grained selection of the vectors to visit.

The advantages of M-LVQ over LVQ do not come for free. First, we need to additionally store $\lceil \log_2 M \rceil$ bits for each vector, as we need to know which center was used to encode the vector in order to compute its similarity to the query. As LVQ-compressed vectors are padded to a multiple of 32 bytes to improve performance~\cite{aguerrebere_similarity_2023}, the additional footprint in M-LVQ can be hidden in practice.

From a computational perspective, M-LVQ is more involved than LVQ. We will use Euclidean distance and inner product as the similarity functions to illustrate our point.

\textbf{Euclidean distance.} Here, during graph traversal, we asses the dissimilarity using $\| \vect{q} - \vect{x}\|_2$. With LVQ, we use the approximation $\vect{x} \approx Q(\vect{x}) + \vect{\mu}$.
We can thus compute $\vect{q}' = \vect{q} - \vect{\mu}$ once for each query and directly conduct the graph traversal using $\| \vect{q}' - Q(\vect{x}) \|_2$. With M-LVQ, we compute $M$ versions $\vect{q}_m = \vect{q} - \vect{\mu}_m$ of the query. For each vector visited during graph traversal, we select which of these queries to use, computing $\| \vect{q}_m - Q(\vect{x}) \|_2$. Keeping $\{ \vect{q}_m \}_{m=1}^M$, using $dM$ floating-point numbers, is possible for moderate values of $M$.

\textbf{Inner product.} Here, during graph traversal, we asses the similarity using $\langle \vect{q}, \vect{x} \rangle$. With LVQ, we can compute $c = \langle \vect{q}, \vect{\mu} \rangle$ once for each query and directly conduct the graph traversal using $\langle \vect{q}, Q(\vect{x}) \rangle + c$ as the similarity (as $c$ is a global constant for each query, we could even use $\langle \vect{q}, Q(\vect{x}) \rangle$ if we are not interested in the actual distance values). With M-LVQ, we compute $M$ values $c_m = \langle \vect{q}, \vect{\mu}_m \rangle$. For each vector visited during graph traversal, we select which of these values to use, computing $\langle \vect{q} - Q(\vect{x}) \rangle + c_m$. Notice that M-LVQ is particularly efficient in this case as the scalar values $\{ c_m \}_{m=1}^M$ can be easily kept in cache, even for relatively large values of $M$.

\subsection{On the impact of LVQ in search performance}
\label{sec:analyzing_lvq}
Most questions about the robustness of LVQ in different similarity search scenarios can be essentially translated to assessing the relationship between the quantization level and search performance. In the case of graph search, a vector reconstruction error increase (decrease) may lead to a slower (faster) search because it increases (reduces) the number of comparisons between the query and the database vectors required to reach a target recall. Or otherwise said, it increases (reduces) the search window size required to achieve the target recall (see \cref{ssec:introduce_lvq}). Therefore, establishing the relationship between the vector reconstruction error and the search window size can help us answer these questions. The same goes for understanding whether any effort to improve LVQ by reducing its reconstruction error may result in a faster search or not. 

We run LVQ for different choices of $B_1$ and $B_2$ and establish an empirical relationship between the quantization level and the search window size required to achieve a target recall (0.9 10-recall@10). We use the open-images-512-1M dataset (see \cref{ssec:datasets_and_procols}) to illustrate the discussion.  

\begin{figure}
  \centering  
  \includegraphics[width=\columnwidth]{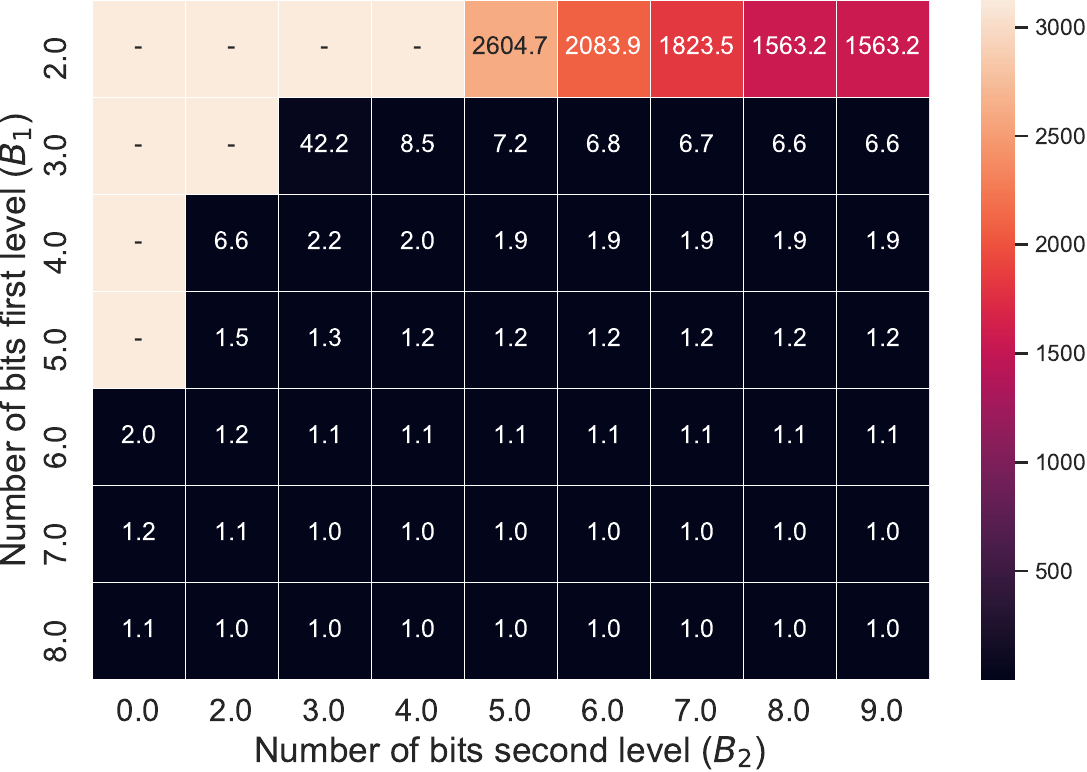}  

  \caption{Heatmap showing the multiplicative increment in search window size (ratio w.r.t using float32 encoded vectors) required to reach 0.9 10 recall@10 for the open-images-512-1M dataset when using LVQ-$B_1\mathsf{x}B_2$ compressed vectors. Different combinations of the number of bits for the first ($B_1$) and second ($B_2$) LVQ-levels represent different noise level scenarios. For large enough $B_1$ and $B_2$ (lower-right corner), there is no search window size increase with respect to using float32 vectors, whereas for small enough $B_1$ and $B_2$ the target recall cannot be reached regardless of the search window size increase (the increment is not reported for these cells). The required window size strongly depends on the reconstruction error in the first level (represented by $B_1$) because there is only so much the re-ranking can fix when having a bad set of neighbor candidates. On the contrary, for a given $B_1$, the quantization error at the second LVQ level only impacts the search window size for very small $B_2$ ($B_2\leq 3$).}
  \label{fig:quant_error_heatmap}
\end{figure}

In \cref{fig:quant_error_heatmap}, we represent the multiplicative increase in search window size needed to achieve 0.9 10-recall@10 when using LVQ-$B_1\mathsf{x}B_2$ versus the one required when using full precision vectors (i.e., float32 encoded vectors). We observe, in the lower-right corner, that for large enough $B_1$ and $B_2$ (that yield a small quantization error), the search window size remains unchanged (i.e., with a multiplicative increase of 1). Surprisingly, with as few as 10 bits ($B_1=8$ and $B_2=2$), LVQ can match the search window size calibrated with the 32-bit floating-point encoding. Notice that Aguerrebere et al.~\cite{aguerrebere_similarity_2023} showed that using float16-encoded vectors leads to a similar search accuracy than float32 and, in this case, LVQ-$8\mathsf{x}2$ leads to a sizeable reduction of 6 fewer bits per dimension, which for $d=512$ and $n=10^9$ leads to saving 357 gigabytes.

For small enough $B_1$ and $B_2$, the target recall cannot be reached regardless of the search window size increase. Interestingly, setting $B_1=2$ can still achieve the target recall (upper-right corner of \cref{fig:quant_error_heatmap}). In such a case, there is a very large increase (> 1500x) in the search window size with respect to using full precision vectors. For particular systems with limited memory, this may be an advantageous setting. 

In \cref{fig:quant_error_heatmap}, we observe that $B_1$ is the critical factor to set the search window size. The value of $B_2$ only affects the final ordering of neighbor candidates obtained using the first-level encoding (with $B_1$ bits) in the graph search. For $B_1 \geq 7$ (bottom rows), the dependency of the window size on $B_2$ is relatively small to almost non-existent, as the original order in the set of neighbor candidates is already accurate and barely needs re-ranking. The behavior changes for $B_1 \leq 5$ (upper rows), where the re-ranking step becomes critical to restore the correct order of the neighbor candidates.

\subsubsection{M-LVQ versus LVQ}
\label{sssec:evaluate_lvq_alternatives}

Let $\varepsilon_1$ be the mean squared quantization error for the first level. Given a dataset, this error is a deterministic function of $B_1$ for LVQ (see \cref{ssec:introduce_lvq}) and for M-LVQ (see \cref{def:mlvq}), given by
\begin{equation}
    \varepsilon_1 (B_1) = \sum_{i=1}^{n} \norm{ \vect{x}_i - \vect{\mu} - Q(\vect{x}_i) }{2}^2 .
    \label{eq:mean_quant_error_first}
\end{equation}

\begin{figure}
  \centering  
  \includegraphics[width=\columnwidth]{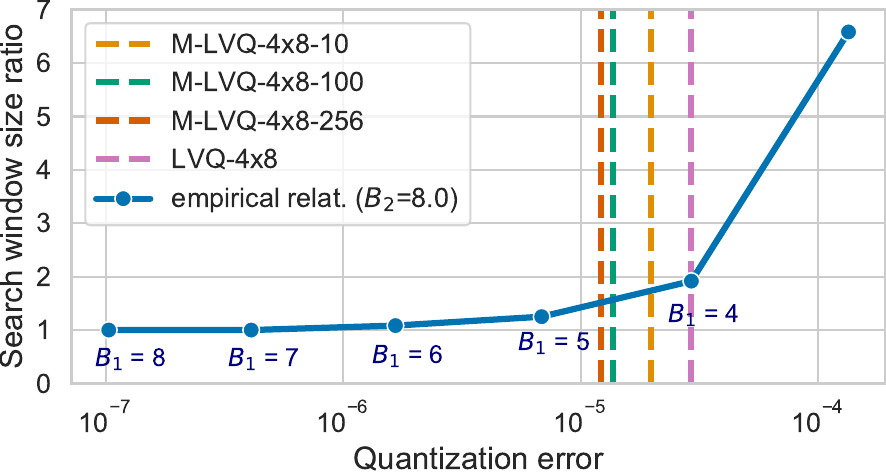}  
  \caption{Search window size ratio (with respect to using float32-encoded vectors) required to reach 0.9 10 recall@10 on open-images-512-1M (\cref{ssec:datasets_and_procols}) as a function of the average first-level quantization error. We use $B_2=8$, which corresponds to the column $B_2=8$ in \cref{fig:quant_error_heatmap}, and replace $B_1$ by its corresponding average quantization error using \cref{eq:mean_quant_error_first} for LVQ and M-LVQ. The vertical lines mark the reconstruction error achieved by LVQ4x8 and M-LVQ4x8 with different number of means, showing that the required search window size can be considerably reduced by M-LVQ if using 10 to 100 means.}
  \label{fig:quant_error_profile}
\end{figure}

As the empirical curves in \cref{fig:quant_error_heatmap} link $B_1$ and $B_2$ to the search window sizes that achieve a target recall, we can now relate search recall to the quantization error. We compute $\varepsilon_1 (B_1)$ for M-LVQ-$4\mathsf{x}8$, with $B_1=4$ and different values of $M$, and use the empirical curves in \cref{fig:quant_error_heatmap} (for $B_2=8$) to estimate the search window size required by M-LVQ-$4\mathsf{x}8$ to achieve the target recall, without the need to run the graph search with this new quantization. As shown in \cref{fig:quant_error_profile}, using M-LVQ-$4\mathsf{x}8$ can significantly reduce the search window size with respect to LVQ-$4\mathsf{x}8$, with reductions of 15\%, 24\% and 26\% for $M=10$, 100 and 256, respectively. Therefore, we can expect to see an improvement in search performance from using M-LVQ on open-images-512-1M.

Interestingly, in \cref{fig:quant_error_profile} and for $M=100$, M-LVQ-$4\mathsf{x}8$ yields a quantization error smaller than LVQ-4x8, roughly equivalent to a hypothetical $B_1=4.5$ bits (using the tools in \cref{ssec:search_different_noise_levels}, we can provide a sub-bit granularity).
This means that the $M=100$ means provide the equivalent of using an additional half bit per dimension: for 512 dimensions, this amounts to saving 32 bytes out of the hypothetical 288 bytes ($512 \cdot 4.5 / 8$), an 11\% reduction.

The diminishing returns of using a larger number of means can be observed in \cref{fig:quant_error_profile}, as setting $M=256$ reduces the quantization error slightly when compared to $M=100$. Finally, no improvement should be expected by using multiple means in the case of LVQ-8, as it already requires the smallest possible search window size (i.e., the ratio with respect to the search window size required when using full-precision vectors is one). 

\begin{figure}
  \centering  
  \includegraphics[width=\columnwidth]{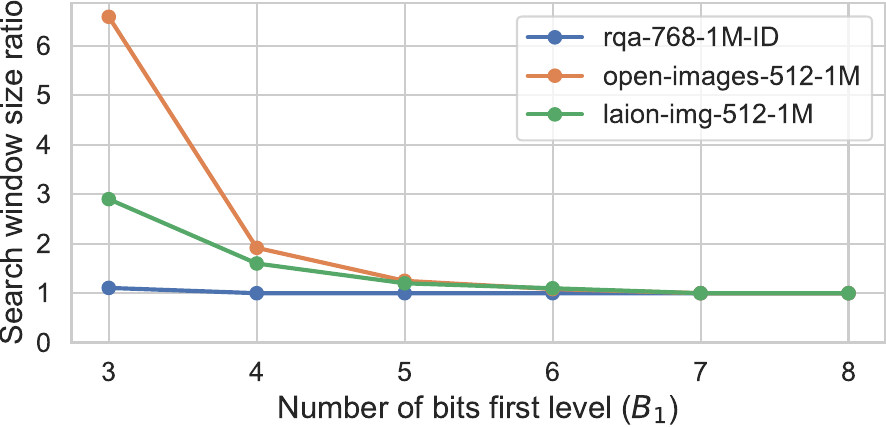}  
  \caption{Search window size ratio (with respect to using float32-encoded vectors) required to reach 0.9 10 recall@10 as a function of the number of bits used for the first quantization level (with $B_2=8$) for various datasets. The behavior varies greatly with the dataset when working on the high quantization error regime (i.e. when using a small number of bits $B_1<5$): open-images-512-1M can achieve a large window size reduction by reducing the quantization error, whereas the effect for rqa-768-1M-ID would be very small. For all datasets, the search window size will not be reduced if going beyond 6 bits.}
  \label{fig:quant_error_all_datasets}
\end{figure}

As shown in \cref{fig:quant_error_all_datasets}, the relationship between the quantization error and the search window size required to achieve the target recall varies greatly across datasets. Unlike open-images-512-1M, the other datasets do not show such a strong dependency between these two variables. Using M-LVQ-$4\mathsf{x}8$ with $M=100$ reduces the search window size with respect to LVQ-$4\mathsf{x}8$, by 0\% (same window size) and 16\% for rqa-768-1M-ID and laion-img-512-1M, respectively. Therefore, no large performance improvements should be expected for these datasets, considering the potential overhead incurred by M-LVQ. \cref{fig:quant_error_all_datasets} also shows that a larger improvement could be expected when using $B_1=3$ bits for the first LVQ level. Regardless, we focus the analysis on $B_1=4$ because, as explained in \cref{ssec:turbo_implementation}, the 4-bit Turbo LVQ implementation can be highly optimized, largely outperforming the performance of 3-bits.

Although the search window size is the main hyperparameter regulating the speed of the graph search, it is not the sole factor. Whether M-LVQ leads to a search performance boost ultimately depends on its computational overhead over LVQ. Nevertheless, analyzing the search window size serves as an excellent proxy for an agile exploration of LVQ variants by enabling the identification of fruitful alternatives without the need of a fully optimized implementation integrated with the graph search. Maybe equally important, this analysis helps understand that reducing the quantization error may not necessarily lead to search performance improvements regardless of their efficiency.

\subsubsection{LVQ for IID streaming data}
\label{sssec:lvq_iid_streaming}
In the streaming case, the index may be initialized with a small subset of vectors thus producing an inaccurate estimate of the mean $\vect{\mu}$. For IID data, the standard error of the mean decreases with $\sqrt{n}$, and thus becomes small even for moderate values of $n$. However, there is no direct link allowing to establish the minimum error such that the graph search performance is not affected. We use the ideas from the previous sections to determine how sensitive LVQ compression is to de-meaning inaccuracies and in particular how this may impact search performance for streaming similarity search.

We compute the first-level mean squared quantization error for LVQ, \cref{eq:mean_quant_error_first}, when using different random sample sizes to compute $\vect{\mu}$: using 1\%, 5\%, 10\%, and 100\% of $\X$. These represent extreme streaming cases where only a very small subset of the vectors is initially available. By using the empirically established relationship between the LVQ error and the search window size in \cref{fig:quant_error_profile}, we confirm these cases exhibit no variation in the search window to achieve 0.9 10-recall@10. Therefore, there will be no search performance degradation even when using as few as 1\% of the vectors to compute $\mu$ 

\section{Experimental Evaluation}
\label{sec:experimental_evaluation}
In this section, we present the results of an exhaustive experimental evaluation supporting our contributions. First, we show that \svs{} establishes the new state-of-the-art in streaming similarity search for the high-throughput high-recall regime. It outperforms its competitors by up to 9.4x and 8.8x in the identically distributed and distribution shift scenarios, respectively. Moreover, we show that this large performance advantage comes with efficient index update operations. Next, we study recall stability over time, another critical aspect of streaming search methods, and show that \svs{} delivers stable recall after long sequences of additions and deletions. Finally, we conduct an ablation study demonstrating the benefits of the novel Turbo LVQ and M-LVQ implementations.

\textbf{Methods.}
For our comparisons, we use the state-of-the-art techniques for streaming similarity search, FreshVamana~\cite{singh_freshdiskann_2021} and HNSWlib~\cite{malkov_efficient_2020}, the de facto standard in the field. This selection only includes the methods that produce state of the art results and have publicly available code. For example, the high-performance streaming techniques in FAISS do not reach an acceptable minimum accuracy. More details about the selection criteria and code availability can be found in \cref{sec:experimental_setup}. For SVS and unless specified, we always use Turbo LVQ. We run experiments with and without M-LVQ and report the best result (see \cref{ssec:ablation_m-lvq} for an ablation study). We report the best out of 5 runs for each method~\cite{aumuller_ann-benchmarks_2020}.
Additional details and hyperparameter selection and tuning are discussed in \cref{sec:experimental_setup}.

\textbf{System Setup.}
We use a 2-socket 3rd generation Intel{\textregistered} Xeon{\textregistered} 8360Y @2.40GHz CPUs with 36 cores and 256GB DDR4 memory (@2933MT/s) per socket running Ubuntu 22.04.\footnote{Performance varies by use, configuration and other factors. Learn more at \url{www.Intel.com/PerformanceIndex}. Performance results are based on testing as of dates shown in configurations and may not reflect all publicly available updates.
No product or component can be absolutely secure. Your costs and results may vary. Intel technologies may require enabled hardware, software or service activation. \textcopyright Intel Corporation.  Intel, the Intel logo, and other Intel marks are trademarks of Intel Corporation or its subsidiaries.  Other names and brands may be claimed as the property of others.}
All experiments are run in a single socket to avoid introducing performance regressions due to remote NUMA memory accesses. We run all methods with and without huge pages and report the best result (see \cref{app:system_setup} for the complete results). More details about the system are available in \cref{sec:experimental_setup}.

\subsection{Datasets and experimental protocols.}
\label{ssec:datasets_and_procols}

We cover a wide variety of scenarios, including different scales ($n=10^6,\dots,10^8$), dimensionalities ($d$=96, 512, 768), and deep learning modalities (texts, images, and multimodal). We also evaluate the IID and distribution shift cases. See the list of the considered datasets in \cref{tab:datasets} and \cref{app:datasets} for additional details.

\begin{table}
   \setlength{\tabcolsep}{3pt}
  \caption{The evaluated datasets, where $n$, $n_q$, and $n_l$ represent the number of database, query, and learning vectors, respecticaly, and $d$ their dimensionality. We generated open-images-512-13M from~\cite{radford_learning_2021, kuznetsova_open_2020} (reproducibility details in ~\cref{suppMat:open-images-dataset} of the supplementary material). In all cases, the original vectors are encoded with float32 values.}
  \label{tab:freq}
  \resizebox{\linewidth}{!}{
  \begin{tabular}{clS[table-format=3]S[scientific-notation=true,print-unity-mantissa=false,table-format=1.1e1]cS[print-zero-exponent=true, 
  print-unity-mantissa=false] S[table-format=4]}
    \toprule
    & Dataset & {$d$} & {$n$} & Similarity & {$n_q$} & {$n_l$}\\
    \midrule
    
    \multirow{4}{*}{\footnotesize \begin{sideways}Small scale\end{sideways}}
     & rqa-768-1M-ID~\cite{tepper_leanvec_2023}   & 768  & 1e6 & inner prod. & \num{1e4} & 5000 \\    
    & open-images-512-1M   & 512  & 1e6 & inner prod. & \num{1e4} & \num{5000} \\    
    & laion-img-512-1M~\cite{schuhmann_laion-400m_2021}   & 512  & 1e6 & inner prod. & \num{1e4} & 5000 \\  
    & deep-96-1M~\cite{babenko_deep_2016}     & 96  & 1e6  & cos sim. & \num{1e4} &  5000 \\     
    \midrule
    
    \multirow{3}{*}{\footnotesize \begin{sideways}Large scale\end{sideways}}
    & rqa-768-10M-ID~\cite{tepper_leanvec_2023}   & 768  & 1e7 & inner prod. & \num{1e4} & 5000 \\ 
    & open-images-512-13M   & 512  & 1.3e7 & inner prod. & \num{1e4} & 5000 \\           
    & deep-96-100M~\cite{babenko_deep_2016}    & 96  & 1e8 & cos sim.  & \num{1e4} & 5000 \\
  \bottomrule
\end{tabular}}
\label{tab:datasets}
\end{table}

We build $\X_0$, $\X_1$, \dots, defined in \cref{ssec:streaming_simsearch}, using the following protocols.

\textbf{Protocol for IID streams.}
We start with setting $\X_0$ as a random sample containing 70\% of the $n$ dataset vectors. At each time $t$, we randomly delete 1\% of the vectors in $\X_{t-1}$ and add 1\% of the vectors in $\X \setminus \X_{t-1}$. For the indices that perform delete consolitations, they are done every every five cycles, i.e., when $t / 5 = 0$.

\textbf{Protocol for streams with distribution shifts.}
Despite the relevance of natural data distribution shifts, there are no standard, openly available, datasets designed to assess the robustness of similarity search methods under such scenarios.\footnote{Baranchuck et al.~\cite{baranchuk_dedrift_2023} recently use two new datasets for this purpose but, to the best of our knowledge, they have not been released publicly by \today.}

To recreate a natural data distribution shift, we introduce a new dataset by generating CLIP~\cite{radford_learning_2021} embedding vectors ($d=512$) for a set of 13M image crops obtained from the Google’s Open Images~\cite{kuznetsova_open_2020} dataset that come with associated class labels (see \cref{suppMat:open-images-dataset} for details). \cref{fig:open-images-dataset} shows examples of the image crops corresponding to four different classes (e.g., box, food, building and fountain), out of the total $F=434$ classes, and the distribution of their vectors. To model the data distribution shift, we initialize the index with the embeddings belonging to a subset of the classes, and then sequentially add the rest of the classes, whose embeddings have diverse distributions (see \cref{fig:open-images-dataset}), thus introducing a distribution shift in the process. 

We also simulate distribution shifts using standard datasets that do not come with class labels by clustering the vectors into $F$ clusters using k-means.

Let $\{\vect{c}_i \in \Real^d \}_{i=1}^F$ be the centroids of the $F$ clusters.
We \textbf{initialize} $\X_0$ with the vectors belonging to the farthest cluster, i.e., $i^* = \argmax_{i} \sum_{j=1}^F  \norm{ \vect{c}_i - \vect{c}_j }{2}^2 $, and those in the clusters closest to $\vect{c}_{i^*}$, up to a total of 5\% of the dataset vectors $n$. This selection ensures that the distribution of the vectors added subsequently is as different as possible from those in $\X_0$ (notice for instance the differences between the \emph{building} and \emph{food} classes in \cref{fig:open-images-dataset}).
Next, there is a \textbf{ramp-up} stage where batches of $B$ vectors of the remaining clusters are added at each $t$ ($B=20$k and $B=200$k for datasets with 1M and over 10M vectors, respectively) until the database contains a large enough number of vectors to make the search problem challenging (i.e., 70\% of the dataset vectors).
Finally, we reach a \textbf{steady-state} where, at each $t$, we perform additions and deletions with batch size $B$ to keep the database size constant. 

Additions are performed by inserting $B$ elements from a cluster chosen at random from the ones reserved for this task. If the chosen cluster contains fewer than $B$ elements, the procedure is repeated until a total of $B$ vectors is reached. Deletions are performed at random. Delete consolidations (see \cref{ssec:graph_construction}) are done every every five cycles ($t / 5 = 0$) for the indices that require them. The query vectors are sampled at each $t$ from a holdout set (to avoid side effects from overlaps) that follows the same distribution as $\X_t$.

With these protocols in hand, we assess the robustness of LVQ to an inaccurate estimate of the sample mean by selecting $\vect{\mu}$ as the mean vector in $\X_0$. In the case of distribution shifts, by selecting $\X_0$ to be as divergent as possible from $\X$, we introduce an extreme distribution shift.

\subsection{Evaluating IID streams}
\subsubsection{LVQ robustness for iid streams}
The analysis presented in \cref{sssec:lvq_iid_streaming} predicts a strong search performance from LVQ even when initialized with a very small subset of vectors. We now proceed to confirm this result and the suitability of LVQ for streaming similarity search with IID data, showing its large performance advantage over the state-of-the-art. The IID setting is evaluated as detailed in \cref{ssec:datasets_and_procols}.

\begin{figure}
  \centering  
  \includegraphics[width=\columnwidth]{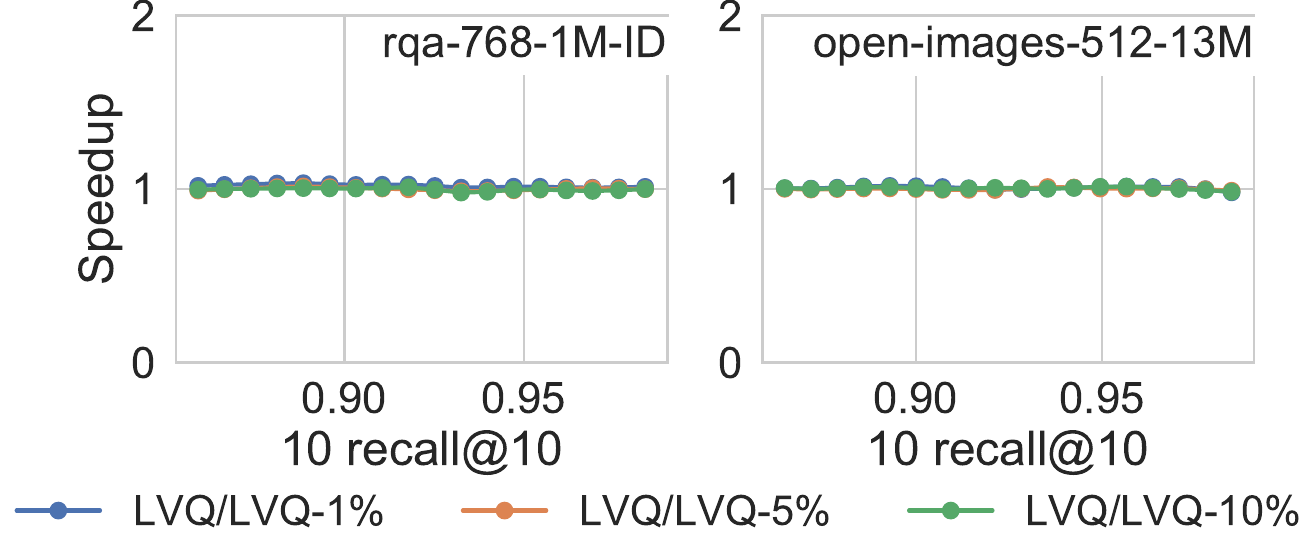}  
  \caption{Similarity search with \svs{} is robust to noisy estimates of the mean. For two datasets and different recall values, we plot the speedup obtained by computing $\vect{\mu}$ from the entire dataset with respect to the throughput obtained when using random samples of 1\%, 5\% and 10\% to compute $\vect{\mu}$. The ratio is 1 for all cases, showcasing LVQ's robustness: computing the mean with as few as 1\% of the vectors is enough to achieve peak performance. The search window size required to achieve each target recall remains unchanged.}
  \label{fig:LVQ_robustness}
\end{figure}

We evaluate the robustness of LVQ to inaccuracies in the mean estimate by building a static graph with full precision vectors and running the search using LVQ-4x8 with $\vect{\mu}$ computed using a random sample containing 1\%, 5\% or 10\% of the vectors in $\X$. The results in \cref{fig:LVQ_robustness} confirm the predictions in \cref{sssec:evaluate_lvq_alternatives}, where the same search performance is achieved when computing $\vect{\mu}$ from the full sample mean or from as few as 1\% of $\X$ (see \cref{fig:LVQ_robustness_appendix} in \cref{app:extra_results} for other datasets).  

\begin{figure*}
  \centering  
  \includegraphics[width=\textwidth]{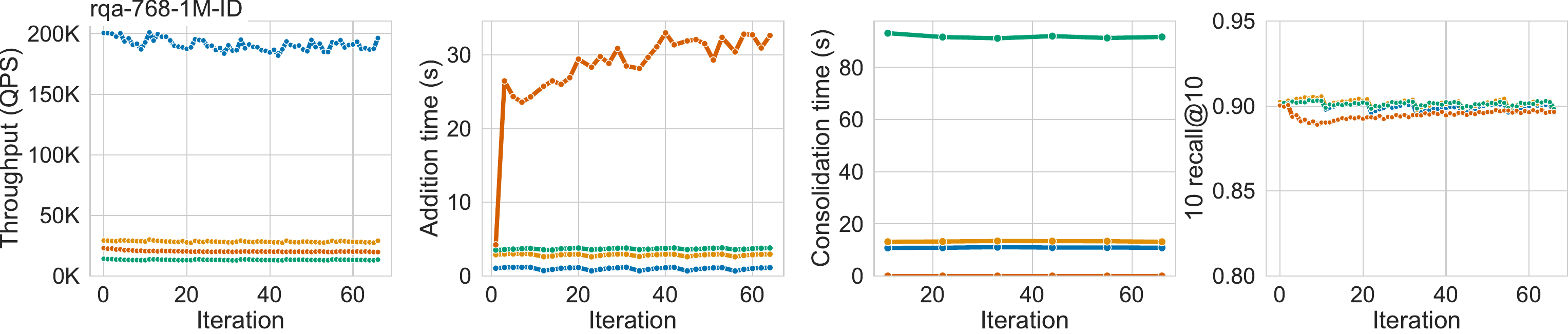}
  
  \includegraphics[width=\textwidth]{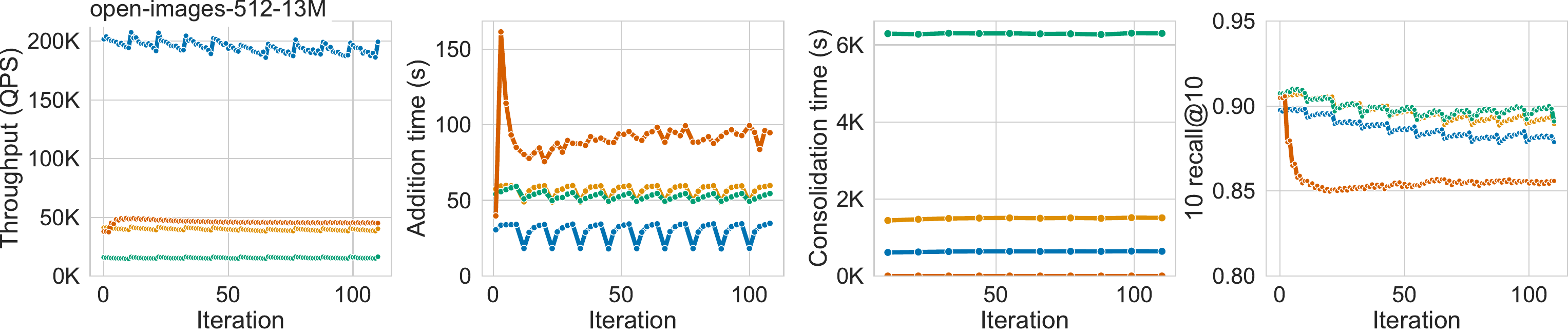}
  
  \includegraphics[width=\textwidth]{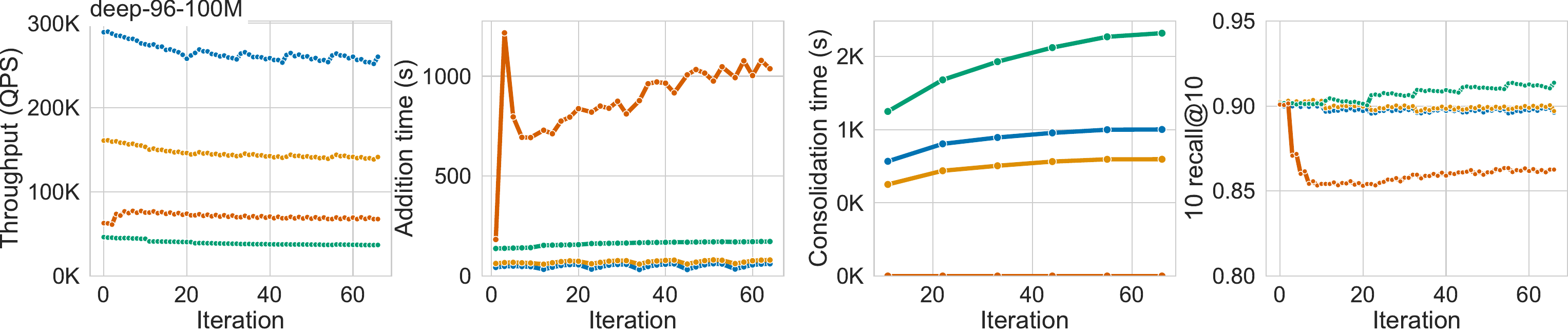}
    \includegraphics[width=0.65\textwidth]{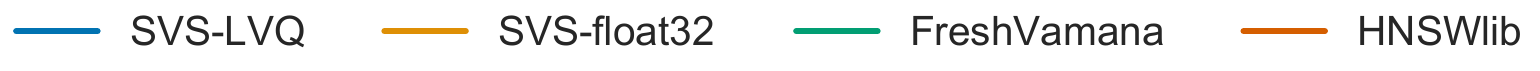}
  \caption{Performance comparisons for IID streaming data (see the protocol in \cref{ssec:datasets_and_procols}). From left to right: search throughput at iso-recall (10-recall@10 of 0.9), runtime for the batched additions, runtime for the delete consolidation rounds, and recall for a fixed search window size (set to achieve 0.9 10-recall@10 at $t=0$). The \svs{} search performance is 9.4x (top), 4.9x (center) and 4.6x (bottom) higher than that of its closest competitor, with a consistent advantage over time. LVQ plays a key role as it gives \svsfullp{} a 6.7x (top), 4.7x (center) and 1.8x (bottom) performance boost with respect to using float32-encoded vectors. \svs{} outperforms FreshVamana in addition runtime by 3.6x (top), 1.8x (center) and 3.2x (bottom) and in delete consolidation runtime by 8.4x (top), 10.0x (center) and 1.6x (bottom). \svs{} is 28.0x (top), 3.0x (center) and 17.7x (bottom) faster than HSNWlib for additions. HNSWlib does not perform delete consolidations (the reported value is zero), causing a degradation in recall over time (rightmost column). \svsfullp{}{} manages to maintain a stable recall over time. A slight initial drop is observed for open-images-512-13M, but recall stabilizes after 60 iterations. See \cref{app:extra_results} for additional datasets.}
  \label{fig:bench_qps_over_time}  
\end{figure*}

\subsubsection{Comparison to other methods}
\label{sssec:bench_iid_other_methods}
\cref{fig:bench_qps_over_time} shows that \svs{} has a large search performance advantage over its competitors, which is sustained over time. Here, LVQ is the main contributor as evidenced by the large performance gap between \svs{} and \svsfullp{}-float32. The results in \cref{tab:benchmarks_qps_ratios} show that \svs{}'s superiority holds both for small and large scale datasets, achieving performance boosts of up to 9.4x and 4.6x, respectively, for all the considered dimensionalities. Moreover, the search performance advantage is accompanied by efficient index update operations. As shown in \cref{tab:benchmarks_qps_ratios}, \svs{}'s addition and consolidation times outclass the competition across datasets (up to 3.6x and 17.7x faster, respectively). HNSWlib supports delete requests by adding them to a blacklist and removing the deleted vectors from the retrieved nearest neighbors. The slots in the delete list will be used for future vectors, but there is not a proper notion of consolidation like the FreshVamana algorithm has. Therefore, the reported consolidation time is zero for HNSWlib. The time taken by deletions is not reported as it is negligible compared to the other tasks for all methods. See \cref{fig:bench_qps_over_time_appendix} in \cref{app:extra_results} for the results with other datasets.

\begin{table*}[t]
  \caption{Summary of the search performance (at 0.9 10 recall@10), time taken by index updates (additions and delete consolidations), and recall (achieved at fixed search window size) for all the considered datasets in the IID (top) and data distribution shift (bottom) streaming scenarios (see the protocols in \cref{ssec:datasets_and_procols}). For each metric, for the IID case we report the average across iterations and for the distribution shift case we report the average for the iterations at steady-state, where additions and deletions are performed while keeping the total number of vectors fixed. The highest QPS and lowest index update times are shaded. Consolidation time is not reported for HNSWlib as it does not perform this task (see \cref{sssec:bench_iid_other_methods}).}
  \label{tab:benchmarks_qps_ratios}

  \sisetup{detect-weight=true,detect-inline-weight=math}
  
  \resizebox{\linewidth}{!}{
  \begin{tabular}
  {
    l %
    l %
    S[table-format=7] %
    S[table-format=6] S[table-format=1.1] %
    S[table-format=6] S[table-format=2.1] %
    S[table-format=6] S[table-format=1.1] %
    S[table-format=1.1] %
    S[table-format=3.1] S[table-format=1.1] %
    S[table-format=4.1] S[table-format=2.1] %
    S[table-format=4.1] %
    S[table-format=4.1] S[table-format=2.1] %
    S[table-format=1.1] S[table-format=1.3] %
  } 
    \toprule
 && \multicolumn{7}{c}{Search performance (QPS)} & \multicolumn{5}{c}{Additions Time (s)} & \multicolumn{3}{c}{Consolidation Time (s)} & \multicolumn{2}{c}{Recall} \\
    \cmidrule(lr){3-9} \cmidrule(lr){10-14} \cmidrule(lr){15-17} \cmidrule(lr){18-19} 
 && {SVS-LVQ} & \multicolumn{2}{c}{SVS-float32} & \multicolumn{2}{c}{FreshVamana} & \multicolumn{2}{c}{HNSWlib} & {SVS-LVQ} & \multicolumn{2}{c}{FreshVamana} & \multicolumn{2}{c}{HNSWlib} & {SVS-LVQ} & \multicolumn{2}{c}{FreshVamana} & \multicolumn{2}{c}{SVS-LVQ} \\
 \cmidrule(lr){3-3} \cmidrule(lr){4-5} \cmidrule(lr){6-7} \cmidrule(lr){8-9} \cmidrule(lr){10-10} \cmidrule(lr){11-12} \cmidrule(lr){13-14} \cmidrule(lr){15-15} \cmidrule(lr){16-17} \cmidrule(lr){18-19}
 && {QPS} & {QPS} & {Ratio} & {QPS} & {Ratio} & {QPS} & Ratio & {Time} & {Time} & {Ratio} & {Time} & {Ratio} & {Time} & {Time} & {Ratio} & {Avg.} & {Std.} \\
\midrule
\multirow{7}{*}{%
\begin{sideways}
    IID%
\end{sideways}}

 & deep-96-1M & \cellcolor[gray]{0.9}1257998 & 700465 & 1.8 & 197460 & 6.4 & 220001 & 5.7 & \cellcolor[gray]{0.9}0.2 & 0.4 & 1.6 & 1.7 & 7.4 & \cellcolor[gray]{0.9}2.4 & 4.8 & 2.0 & 0.9 & 0.003\\
 & open-images-512-1M & \cellcolor[gray]{0.9}449221 & 96846 & 4.6 & 36202 & 12.4 & 69412 & 6.5 & \cellcolor[gray]{0.9}0.7 & 1.9 & 2.9 & 2.9 & 4.4 & \cellcolor[gray]{0.9}7.5 & 55.3 & 7.4 & 0.9 & 0.003\\
 & laion-img-512-1M & \cellcolor[gray]{0.9}629631 & 122849 & 5.1 & 43693 & 14.4 & 85717 & 7.3 & \cellcolor[gray]{0.9}0.5 & 1.6 & 3.2 & 3.6 & 7.3 & \cellcolor[gray]{0.9}4.8 & 36.4 & 7.6 & 0.9 & 0.002\\
 & rqa-768-1M-ID & \cellcolor[gray]{0.9}190975 & 28409 & 6.7 & 13361 & 14.3 & 20411 & 9.4 & \cellcolor[gray]{0.9}1.0 & 3.7 & 3.6 & 28.4 & 28.0 & \cellcolor[gray]{0.9}10.9 & 91.8 & 8.4 & 0.9 & 0.002\\
 & open-images-512-13M & \cellcolor[gray]{0.9}175891 & 37509 & 4.7 & 15053 & 11.7 & 35830 & 4.9 & \cellcolor[gray]{0.9}29.9 & 53.4 & 1.8 & 89.9 & 3.0 & \cellcolor[gray]{0.9}636.2 & 6345.9 & 10.0 & 0.9 & 0.006\\
 & rqa-768-10M-ID & \cellcolor[gray]{0.9}98985 & 16312 & 6.1 & 6807 & 14.5 & 11982 & 8.3 & \cellcolor[gray]{0.9}32.7 & 75.1 & 2.3 & 836.1 & 25.6 & \cellcolor[gray]{0.9}543.8 & 9647.6 & 17.7 & 0.9 & 0.002\\
 & deep-96-100M & \cellcolor[gray]{0.9}253319 & 140606 & 1.8 & 40633 & 6.2 & 55347 & 4.6 & \cellcolor[gray]{0.9}50.0 & 161.6 & 3.2 & 886.3 & 17.7 & \cellcolor[gray]{0.9}935.2 & 1463.6 & 1.6 & 0.9 & 0.001\\
\midrule
\multirow{4}{*}{%
\begin{sideways}
    Dist.~Shift%
\end{sideways}}
& open-images-512-1M & \cellcolor[gray]{0.9}474312 & 92994 & 5.1 & 34892 & 13.6 & 73239 & 6.5 & 4.1 & \cellcolor[gray]{0.9}2.9 & 0.7 & 5.4 & 1.3 & \cellcolor[gray]{0.9}3.3 & 15.5 & 4.8 & 0.9 & 0.002\\
 & rqa-768-1M-ID & \cellcolor[gray]{0.9}197474 & 30310 & 6.5 & 14264 & 13.8 & 22525 & 8.8 & \cellcolor[gray]{0.9}3.5 & 6.8 & 1.9 & 50.5 & 14.3 & \cellcolor[gray]{0.9}3.3 & 34.9 & 10.6 & 0.9 & 0.003\\
 & open-images-512-13M & \cellcolor[gray]{0.9}173529 & 34859 & 5.0 & 15067 & 11.5 & 35952 & 4.8 & 120.6 & \cellcolor[gray]{0.9}84.3 & 0.7 & 140.4 & 1.2 & \cellcolor[gray]{0.9}196.0 & 1311.4 & 6.7 & 0.9 & 0.004\\
 & rqa-768-10M-ID & \cellcolor[gray]{0.9}103910 & 16934 & 6.1 & 7244 & 14.3 & 12871 & 8.1 & \cellcolor[gray]{0.9}63.0 & 153.5 & 2.4 & 1574.4 & 25.0 & \cellcolor[gray]{0.9}134.2 & 2156.3 & 16.1 & 0.9 & 0.002\\
    \bottomrule
  \end{tabular}}
\end{table*}

\subsection{Evaluating streams with distribution shifts}

We now evaluate LVQ under data distribution shifts, confirming its suitability for dynamic indexing under this challenging scenario. Moreover, we show the large advantage of \svs{} over competitive alternatives, thus establishing the new state-of-the-art for streaming similarity search under distribution shifts. 
The data distribution shift is evaluated as detailed in \cref{ssec:datasets_and_procols}.

\subsubsection{LVQ robustness under data distribution shifts} 
\begin{figure}
  \centering  
  \includegraphics[width=\columnwidth]{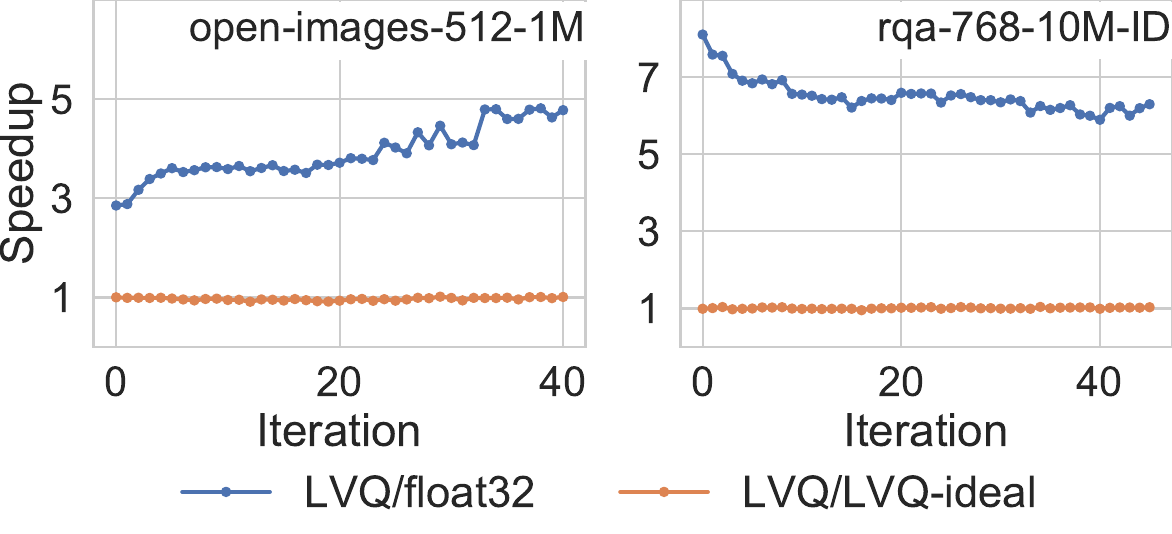}
  \caption{\svs{} achieves significant speedups over SVS-float32 (blue curve) under data distribution shifts (see the protocol in \cref{ssec:datasets_and_procols}). When compared to SVS-LVQ-ideal, which represents the ideal and impractical scenario where the entire database is re-encoded at each $t$ using LVQ with the sample mean from $\X_t$, the constant speedup of 1 in the orange curve indicates no degradation in search performance.}
  \label{fig:LVQ_robustness_distro_shift}
\end{figure}

We compare the search performance of \svs{} to: (i) \svsfullp{} using full-precision vectors (float32); and (ii) the ideal scenario where, after each round of additions and deletions, all the vectors are re-encoded using LVQ with the sample mean from $\X_t$. We call this setting SVS-LVQ-ideal because updating $\vect{\mu}$ at each iteration is a best-case scenario for LVQ. This is practically unrealistic as re-encoding all vectors is computationally expensive and requires keeping the full precision vectors in main memory. As observed in \cref{fig:LVQ_robustness_distro_shift}, the performance of \svs{} remains high under data distribution shifts, achieving the same performance as SVS-LVQ-ideal and maintaining over time its advantage over \svsfullp{} with full-precision vectors holds for all the evaluated datasets (see \cref{fig:LVQ_robustness_distro_shift_appendix} in \cref{app:extra_results} for other datasets).

\subsubsection{Comparison to other methods}
\begin{figure*}
  \centering    
  \includegraphics[width=\textwidth]{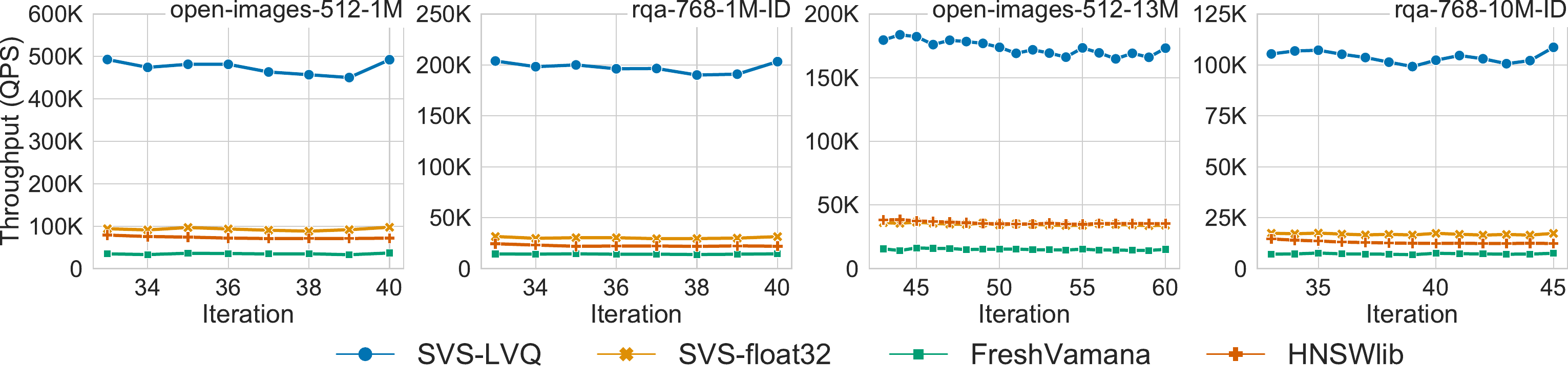}    
  \caption{Search performance under data distribution shift at 0.9 10-recall@10 during the steady-state phase, where additions and deletions are performed while keeping the total number of vectors fixed (see the protocol in \cref{ssec:datasets_and_procols}).  \svs{} offers a large performance advantage (of up to 8.8x, see \cref{tab:benchmarks_qps_ratios}) over HNSWlib that remains consistent over time. LVQ is a key component of this advantage, giving up to a 6.5x performance boost over \svsfullp{}-float32.} 
  \label{fig:bench_qps_over_time_distro_shift}
\end{figure*}

\svs{} exhibits a sizeable search performance advantage even under distribution shifts, outperforming its closest competitor by up to 8.8x and 4.8x at small and large scale, respectively (see \cref{fig:bench_qps_over_time_distro_shift} and \cref{tab:benchmarks_qps_ratios}). \svs{}'s edge consistently holds over time. The competitive advantage of \svsfullp{} is largely due to LVQ as \svs{} outperforms \svsfullp{}-float32 by up to 6.5x. LVQ is robust to an unfavorable initialization, being able to maintain the search performance advantage under the data distribution shift over time. \svs{} also has the lead in additions and delete consolidations, except for the open-images-512-1M and open-images-512-13M where FreshVamana is 30\% faster for additions, see \cref{tab:benchmarks_qps_ratios}.

\subsection{Recall stability}
An important aspect of indexing methods for streaming similarity search is how stable their search recall is over time. Ideally we would want the recall to stay almost unchanged, and compensate slight decreases by increasing the search window size, hopefully with a negligible hit in search performance. The FreshVamana algorithm~\cite{singh_freshdiskann_2021} (the one implemented in \svsfullp{}{}) was designed to fulfill this requirement and was evaluated for a data stream with identically distributed vectors. Here, we extend the evaluation using SVS to the IID and distribution shift settings. 

In the IID case, we set the search window size at $t=0$ to reach the target recall (0.9 10-recall@10) and keep it fixed thereafter. The results in \cref{fig:bench_qps_over_time} (last column) and \cref{tab:benchmarks_qps_ratios} show that \svsfullp{}{} manages to maintain a stable recall over time for most datasets. For the open-images-512-13M dataset, we observe a slight recall degradation that stabilizes after 60 iterations. 

For streams wih distribution shifts, the search window size is set right after the ramp-up stage and it is kept fixed thereafter. The results in \cref{tab:benchmarks_qps_ratios} confirm that \svs{} maintains a stable recall over time even under data distribution shifts for all the evaluated datasets.

\subsection{Ablation study: Turbo LVQ}
\label{ssec:ablation_turbo_lvq}

The novel Turbo LVQ implementation reduces the number of instructions required to unpack the data. In the following, we evaluate how this impacts distance computation time and overall search performance both for static and dynamic indices.

\begin{table}
\caption{Turbo LVQ is faster (in nanoseconds per distance computations for 4-bit encodings) than the original LVQ when linearly scanning (using a single thread) 500 vectors, sampled from multidimensional normal distributions of varying dimensions. Turbo achieves a boost of up to 25\%. }

\label{tab:turbo} 

  \resizebox{\linewidth}{!}{
  \begin{tabular}{l S[table-format=3.2] *{6}{S[table-format=2.2]}}
    \toprule
 Dims ($d$) & {64} & {128} & {160} & {256} & {512} & {768} & {1024} \\
 \midrule
Turbo LVQ & 6.4 & 9.7 & 11.4 & 16.9 & 33.5 & 50.3 & 64.7 \\
LVQ & 8.2 & 12.9 & 15.0 & 22.7 & 43.5 & 64.5 & 85.3 \\ 
\midrule
Speedup (\%) & 22.4 & 24.6 & 24.3 & 25.6 & 23.1 & 22.0 & 24.2 \\
\bottomrule
\end{tabular}}
\end{table}

We conduct an experiment to show the speedup in distance computations of 4-bit Turbo LVQ over the original 4-bit LVQ. First, we construct a random 500 vector dataset with a set number of dimensions using a unit normal distribution for each dimension. Next, we generate a query vector using the same process and record how long it takes to compute the inner product between the query vector and every element in the dataset. Choosing a small dataset size (500 vectors) helps mitigate the overhead of memory bandwidth, to avoid measuring how fast data is read from memory rather than the speed of the computation kernel. We perform and time the operation $10^5$ times to reduce the noise in the measurements. \cref{tab:turbo} shows the time per distance-computation (in nanoseconds) for the original LVQ and Turbo-LVQ, which is 24\% faster on average.

\begin{figure}
  \centering  
  \includegraphics[width=\columnwidth]{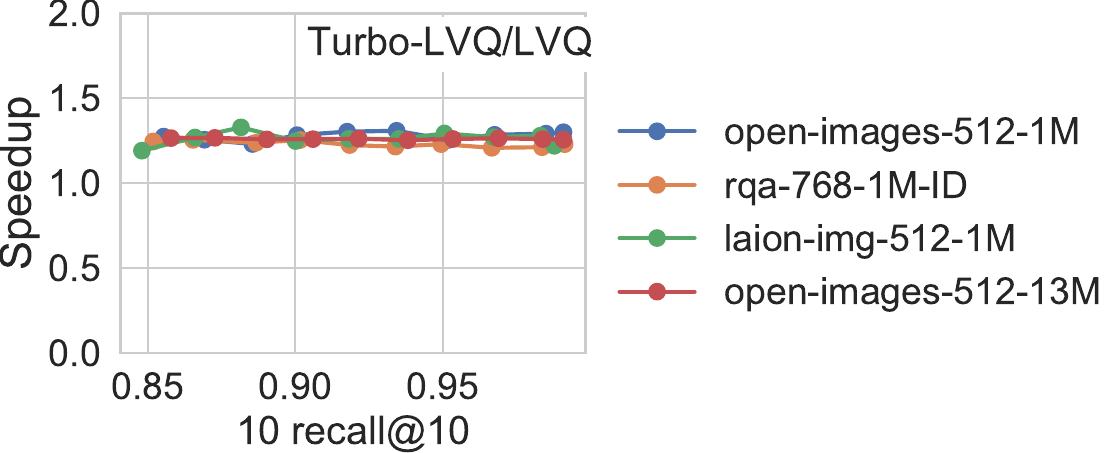} 
  \caption{For static indexing, Turbo LVQ speeds up the graph search compared to the original LVQ at different accuracies. Turbo LVQ-4x8 outperforms Sequential LVQ-4x8 by up to 28\% and 26\% for small and large scale datasets, respectively.}
  \label{fig:turbo_lvq_ablation}
\end{figure}  

\begin{figure}
  \centering  
  \includegraphics[width=\columnwidth]{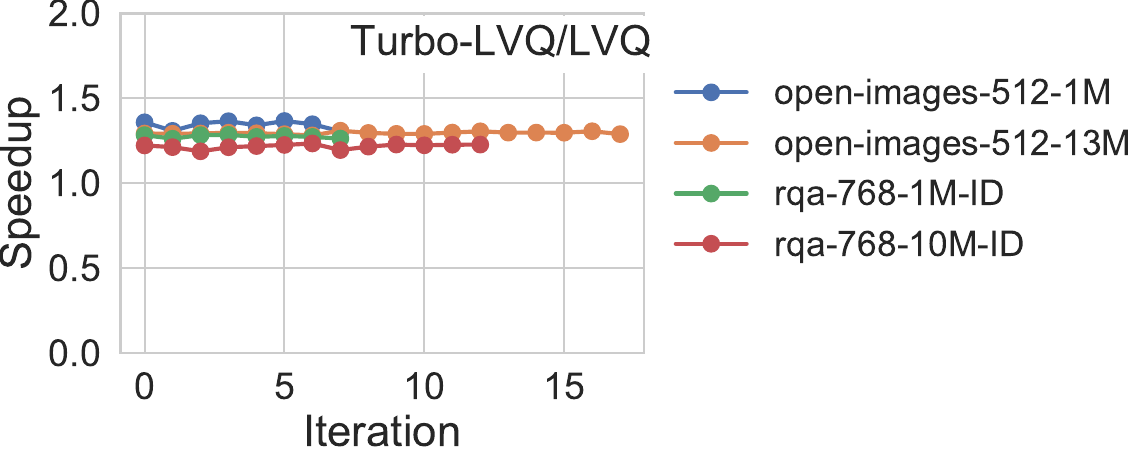} 
  \caption{For streaming-data, Turbo LVQ-4x8 speeds up search performance over the original LVQ-4x8 during the steady-state (see the protocol in \cref{ssec:datasets_and_procols}). Turbo LVQ has a consistent advantage of 34\%, 29\%, 27\% and 22\%, for open-images-512-1M, open-images-512-13M, rqa-768-1M-ID, and rqa-768-10M-ID, respectively. Iteration 0 here corresponds to the beginning of the steady-state phase, which have different duration for different datasets.}
  \label{fig:turbo_lvq_ablation_distroShift}
\end{figure}

We evaluate the differences in search performance between the original and Turbo LVQ for static indices in \cref{fig:turbo_lvq_ablation}. We build the graph with full precision vectors and run the search with vectors encoded using the corresponding LVQ setting. \svsfullp{} powered with Turbo LVQ-4x8 outperforms the original LVQ-4x8 by up to 28\% and 26\% for small and large scale datasets, respectively. In this experiment, we consider the datasets for which LVQ-4x8 is the best setting. As shown in \cref{fig:turbo_lvq_ablation_distroShift}, similar results are obtained for dynamic indices, where Turbo LVQ-4x8 achieves a boost of up to 34\% over vanilla LVQ-4x8. 

\subsection{Ablation study: Multi-Means LVQ}
\label{ssec:ablation_m-lvq}

\begin{figure}
  \centering  
  
  \includegraphics[width=\columnwidth]{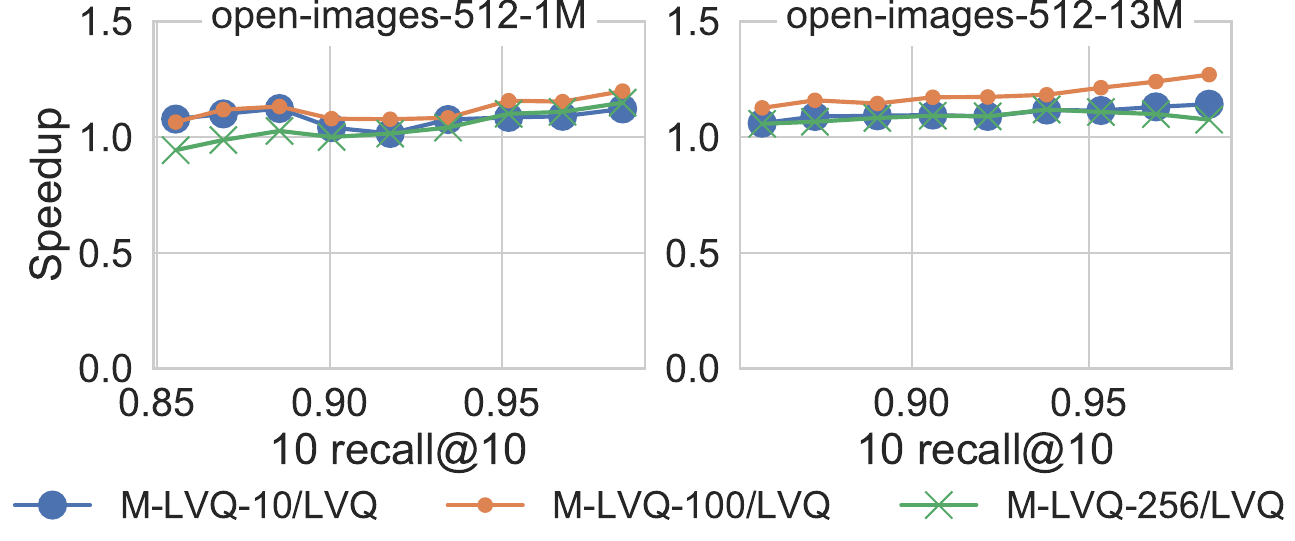}  
  
  \caption{Search performance speedup achieved with M-LVQ over LVQ for the open-images-512 dataset (1M and 13M) using $M=10,100,256$ for static indexing ($B_1=4$, $B_2=8$). The combination of improved compression accuracy and efficient management of the overhead incurred by having extra means leans on the positive side for this dataset. For 13M vectors, it gives a performance boost ranging from 12\% to 27\%, for increasing search accuracy (0.85 to 0.98 10 recall@10), using $M=100$. For 1M vectors the boost ranges from 6\% to 20\%. Using $M=100$ gives a larger speedup than $M=256$, showing that increasing $M$ may eventually hurt performance. 
  }
  \label{fig:LVQ_M-LVQ}
\end{figure}

The analysis presented in \cref{sssec:evaluate_lvq_alternatives} serves as a diagnosis tool to quickly determine at which combinations of $B_1$ and $B_2$, we may see performance benefits from using multiple means for a given dataset. Ultimately, this determination does not depend solely on the M-LVQ technique, but it is heavily tied to the speed of its implementation. In this section, we assess the merits of M-LVQ over LVQ in the context of the state-of-the-art graph search in SVS for static and dynamic indexing, which confirm its superiority in certain cases (see \cref{app:mlvq_implementation} for details).

For the static indexing comparison, we build the graph with full-precision vectors and run searches using LVQ and M-LVQ with $M=10, 100, 256$, where in all cases $B_1=4$, $B_2=8$. The results in \cref{fig:LVQ_M-LVQ} are consistent with the prediction of our previous analysis based on the noise model for LVQ (\cref{sssec:evaluate_lvq_alternatives}). From the evaluated datasets, open-images-512 is the one with the largest sensitivity to the search window size and it is therefore the one showing a performance advantage for M-LVQ, of 8\% and 17\% (at 0.9 10 recall@10) for the 1M and 13M scales, respectively (see \cref{fig:LVQ_M-LVQ}). For the rest of the datasets, there is none to a slight performance improvement but, interestingly, the overhead of using up to 100 means hardly affects performance (see \cref{fig:LVQ_M-LVQ_appendix} in \cref{app:extra_results}).

\begin{figure}
  \centering  
  \includegraphics[width=\columnwidth]{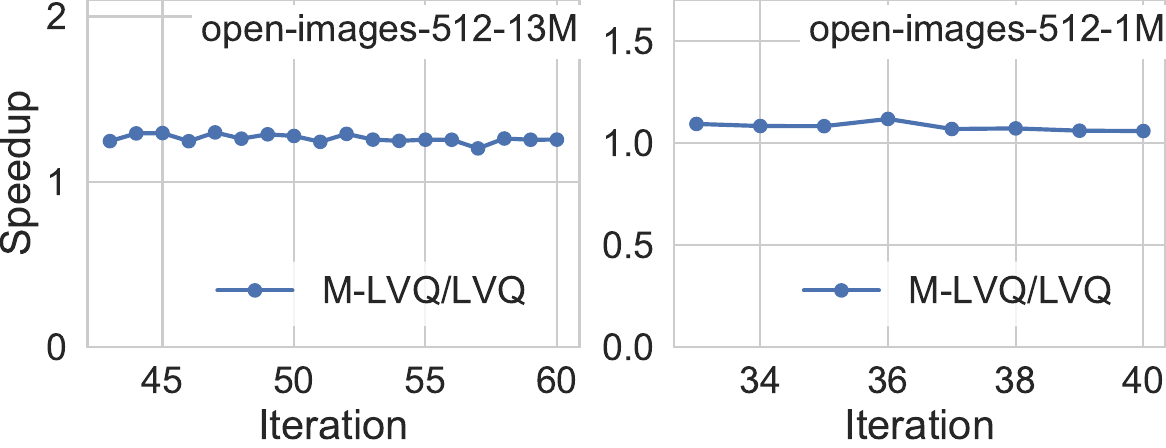} 
  \caption{Search performance speedup obtained by M-LVQ over LVQ for the open-images-512 dataset (1M and 13M) using $M=100$ at 0.9 10 recall@10 accuracy during the steady-state phase, where additions and deletions are performed while keeping the total number of vectors fixed (see the protocol in \cref{ssec:datasets_and_procols}). M-LVQ achieves 8\% and 25\% higher QPS than LVQ for 1M and 13M vectors, respectively. }
  \label{fig:LVQ_MLVQ_dynamic}
\end{figure}

\begin{figure}
  \centering  
  \includegraphics[width=\columnwidth]{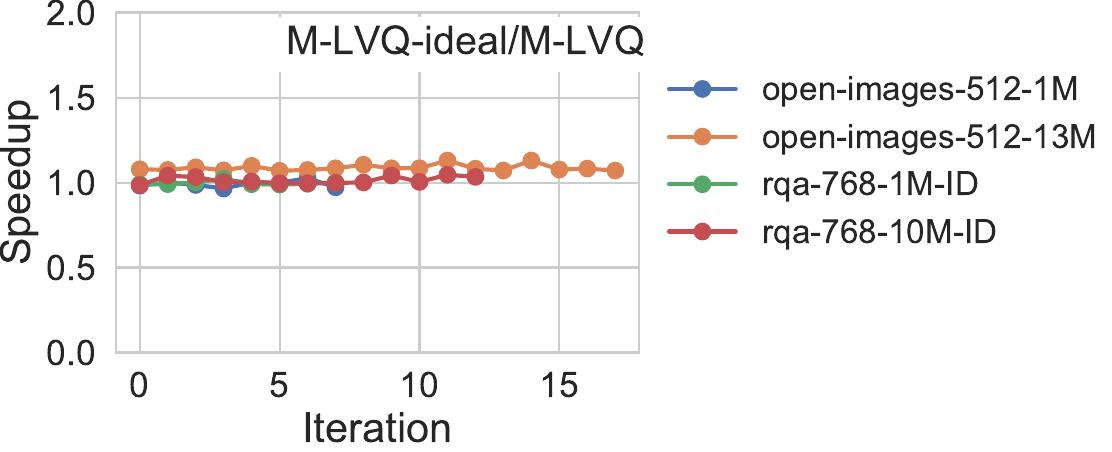} 
  \caption{Search performance speedup obtained by M-LVQ-ideal over M-LVQ during the steady-state phase, where additions and deletions are performed while keeping the total number of vectors fixed (see the protocol in \cref{ssec:datasets_and_procols}). In M-LVQ-ideal, we update the $M$ means at each time $t$ by running k-means on $\X_t$, yielding an ideal but impractical baseline. In all cases, there is no meaningful advantage in updating the means over time (up to 9\% on open-images-512-13M after ramp-up), which would involve complex and costly procedures. Iteration 0 here corresponds to the beginning of the steady-state phase, which have different duration for different datasets.}
  \label{fig:LVQ_MLVQ_dynamic_noNeedUpdate}
\end{figure}

For the comparison on streaming data, we follow the protocol described in \cref{ssec:datasets_and_procols}, using $B_1=4$, $B_2=8$, and $M=100$ for M-LVQ. Similarly to the static case, a performance advantage is observed for the open-images-512 dataset, with 8\% and 25\% boosts for the 1M and 13M scales respectively (see \cref{fig:LVQ_MLVQ_dynamic}). No improvement is observed for other datasets (see \cref{fig:LVQ_MLVQ_dynamic_appendix} in \cref{app:extra_results}).

M-LVQ is better suited than LVQ for updating the encoded vectors dynamically over time. With M-LVQ, it is possible to only update the vectors corresponding to the centers that are shifting, avoiding to entirely re-encode the database. As in LVQ, the practical advantages of carrying this dynamic re-encoding will depend on the characteristics of the stream. To carry this evaluation, we create an idealized M-LVQ variant (M-LVQ-ideal) where, at each time $t$, the entire dataset is re-encoded using centers computed from scratch by running k-means on $\X_t$. The optimality of M-LVQ-ideal, although practically unfeasible due to its extreme computational cost, allows to gauge the potential improvements obtained by updating the centers. The results in \cref{fig:LVQ_MLVQ_dynamic_noNeedUpdate} show that, for most datasets, there is no advantage in updating the centers over time. A performance improvement of 9\% is observed in open-images-512-13M, suggesting that updating the means may bring benefits at larger scales. The introduction of an algorithm for online M-LVQ encoding, e.g., following ideas from k-means for evolving data streams~\cite{bidaurrazaga_k-means_2021}, is left for future work.

\section{Related Work}
\label{sec:related_work}

Among the vast variety of similarity search algorithms, some can be naturally extended to the streaming case whereas others need to be adapted to this scenario. Locality sensitive hashing (LSH)~\cite{sundaram_streaming_2013,wang_survey_2018,jafari_survey_2021} based methods come in both flavours. Vector additions and deletions are trivial for the data-independent LSH-based methods~\cite{datar_locality-sensitive_2004,gionis_similarity_1999}, whereas alternatives have been proposed to adapt the data-dependent LSH-based approaches to the streaming case~\cite{zhang_continuously_2020}. Nevertheless, these are not the preferred solutions for the high-accuracy and high-performance regime as they struggle to simultaneously achieve both. 

Trees~\cite{cayton_fast_2008,muja_scalable_2014,silpa-anan_optimised_2008} are another classic approach in similarity search, but they do not scale well with dimensionality and they are largely outperformed by other methods in the static indexing scenario. Inverted indices combined with data quantization are a widely adopted solution for static indexing in settings with limited memory capacity. Baranchuck et al.~\cite{baranchuk_dedrift_2023} study IVF indices for streaming similarity search under data distribution shifts, showing that they are largely affected if not updated accordingly, and propose algorithms to handle this scenario. In contrast, graph-based methods, as discussed next, organically handle online updates of the graph.

In this work we focus on graph-based methods~\cite{fu_fast_2019,malkov_efficient_2020,subramanya_diskann_2019} because they are the state-of-the-art for static~\cite{aguerrebere_similarity_2023} and dynamic indexing~\cite{singh_freshdiskann_2021}, showing excellent search performance at very high accuracy. Moreover, graph-based approaches have shown to scale well with the number of vectors in the database~\cite{aguerrebere_similarity_2023} as well as with the vector's dimensionality~\cite{tepper_leanvec_2023}. The state-of-the-art graph-based indexing algorithm FreshVamana~\cite{singh_freshdiskann_2021} supports additions and deletions while maintaining a stable search accuracy over long streams of updates. HNSWlib~\cite{malkov_efficient_2020}, one of the most widely adopted graph-based approaches, also supports index updates. %
In both cases, these updates come as natural extensions of the graph construction process (e.g., see \cref{ssec:graph_construction}).

Vector compression is a key component of many similarity search solutions. The practical relevance of the streaming case, motivated the adaptation of classical compression techniques for this scenario. Online hashing techniques~\cite{huang_online_2018,cakir_mihash_2017,li_online_2022} apply different strategies to decide whether the hashing function needs to be updated with the new incoming data. Online PQ~\cite{xu_online_2018}, online optimized PQ~\cite{liu_online_2020} and online additive quantization~\cite{liu_online_2021} focus on efficiently updating the quantization parameters (i.e., codebooks) online as new data arrives. They introduce error bounds to theoretically guarantee the accuracy of the online algorithms under this regime. However, these works do not address the bookkeeping necessary to update the encoded database vectors with the evolving quantization parameters

Even if these approaches achieve a higher search recall compared to keeping the quantization parameters fixed over time, the accuracy gap is not very large~\cite{xu_online_2018,liu_online_2020}. Similarly, Baranchuck et al.~\cite{baranchuk_dedrift_2023} showed that the degradation of PQ compression under data distribution shifts has a mild impact on search accuracy. For most datasets, PQ requires a final re-ranking step with full precision vectors to achieve high recall values. We argue that, after re-ranking, this accuracy gap will disappear almost completely. In this work, we observe a similar behavior for LVQ, showing its robustness under data distribution shifts. The computational efficiency of LVQ, combined with its robustness, provide the advantage to outperform PQ for graph-based similarity search~\cite{aguerrebere_similarity_2023}.

\section{Conclusions}
\label{sec:conclusions}
We presented an extensive evaluation of the recently introduced LVQ technique in the context of streaming similarity search, showing that it leads to state-of-the-art search and index modification performance for different dataset chracteristics. To reduce the compression error, we extended LVQ to use multiple means, achieving significant dataset-dependent search performance boosts of up to 27\%, and analyzed when such improvements in vector compression translate to an increase in search accuracy. We also introduced Turbo LVQ, a novel low level implementation of LVQ that brings consistent speedups of up to 28\% for search performance compared to traditional LVQ. Our contributions have been open-sourced as part of SVS.
The open-source Scalable Vector Search library, powered by LVQ and our contributions (which we release publicly), outperforms its closest competitor by up to 9.4x and 8.8x in the IID and data distribution shift cases, respectively.

For future work, we plan on studying other kinds of distribution shifts (e.g., with additions and deletions following a particular temporal pattern, with more extreme variations in the dynamic range of the embeddings), and on proposing an online algorithm to update the multiple LVQ means over time if it becomes necessary under extreme data distribution shifts. Moreover, we will investigate what makes the open-images-512 dataset particularly suitable for M-LVQ, seeking to extend this advantage to other scenarios.

\clearpage

\bibliographystyle{ACM-Reference-Format}
\bibliography{VLDB2024}

\clearpage

\appendix
\onecolumn
\setcounter{page}{1}

\begin{center}
\Huge{Supplementary material\\Locally-adaptive Quantization for Streaming Similarity Search}

\vspace{2em}

\end{center}

\section{Graph-based streaming similarity search} 
\label{app:graph_streaming_sim_search}
Graph-based methods have shown to be the state-of-the-art for \emph{static} indices, enabling fast and highly accurate similarity search~\cite{aguerrebere_similarity_2023}. The key idea is to build a proximity graph, where two nodes are connected if they fulfill a defined neighborhood criterion, and use a best-first traversal to find the nearest neighbor. Proximity graphs serve as approximations of the computationally expensive Delaunay graph, where a best-first traversal is guaranteed to converge to the nearest neighbor. The most widely used proximity graphs are variants of the relative neighborhood graph (RNG) and the k-nearest neighbor graph (kNNG), obtained by using different edge selection strategies~\cite{fu_fast_2019}, adding a hierarchy~\cite{malkov_efficient_2020}, iterating over the dataset multiple times~\cite{subramanya_diskann_2019}, among others. Most of these approaches do not support index updates, so they cannot be used in the streaming case. Singh et al.~\cite{singh_freshdiskann_2021} introduced FreshVamana, a variation of the Vamana~\cite{subramanya_diskann_2019} algorithm that supports additions and deletions without compromising search performance and accuracy over time. In this work, we use the FreshVamana algorithm for its strong and stable performance for streaming similarity search, but our results apply to other graphs-based methods. 

In the following, let $\Gr=(\V,\E)$ be a directed graph with vertices $\V$ corresponding to vectors in a dataset $\X$ and edges $\E$ representing neighbor-relationships between vectors. We denote with $\set{N}(\vect{x})$ the set of out-neighbors of $\vect{x}$ in $\Gr$.

\subsection{Graph search} 
\label{sssec:graph_search}
Graph search involves traversing $\Gr$ using a modified best-first search approach (see pseudo-code in \cref{alg:search}) to retrieve the $k$ approximate nearest vectors to query $\q$ with respect to the similarity function $\simfun$. The search window size \searchWin{} is the parameter setting the trade-off between search accuracy and performance, as increasing \searchWin{} improves the accuracy of the retrieved neighbors by exploring more of the graph thus increasing search time. 

\subsection{Graph construction} 
\label{ssec:graph_construction}
Graph construction involves building a navigable graph with the initial set of vectors, and performing additions and deletions to keep it updated over time. To this end, we follow the approach by Singh et al.~\cite{singh_freshdiskann_2021}.

\paragraph{Initial build} We start at $t=0$ with an initialized graph $\Gr = (V, \emptyset)$ and target maximum out degree $\maxOutDeg{}$, and perform an update routine for each node $\vect{x} \in \X_t$. For this, we retrieve the \searchWin{} nearest neighbors of $\vect{x}$ by running \cref{alg:search} on the current $\Gr$ with $\searchWin{} > \maxOutDeg{}$, and then prune these neighbors by applying \cref{alg:pruning} to obtain the new outward adjacency list $\set{N}(\vect{x})$ for $\vect{x}$ in $\Gr$. Backward edges $(\vect{x},\vect{x}')$ are then added for all $\vect{x}' \in \set{N}(\vect{x})$, and the pruning rule in \cref{alg:pruning} is applied to the maximum degree $\maxOutDeg{}$ if needed. This procedure is done twice through the dataset, one with the pruning relaxation factor $\alpha=1.0$ and the other with $\alpha=1.2$ or $\alpha=0.95$ when using Euclidean distance and inner product, respectively.

\paragraph{Insertions} To insert a new vector $\vect{x}$ in $\Gr$ we follow the same update routine described for the initial build, except that it is run only once. 

\paragraph{Deletions} When a vector $\vect{x}$ is deleted, edges $(\vect{x'},\vect{x''})$ are added if $\vect{x}$ was in the path between them, i.e., if directed edges $(\vect{x'},\vect{x})$ and $(\vect{x},\vect{x''})$ existed in $\Gr$. Then, if $\set{N}(\vect{x'}) > \maxOutDeg$, the pruning rule in \cref{alg:pruning} is applied to $\set{N}(\vect{x'})$. 

The pruning relaxation factor $\alpha$ is set to retain density of the modified graph during both insertions and deletions ($\alpha=1.2$ or $\alpha=0.95$ when using Euclidean distance and inner product, respectively).

\paragraph{Delete consolidation} Deletions are performed in a lazy fashion to avoid an excessive compute overhead. When a vector is deleted, it is added to a list of deleted elements but not immediately removed from $\Gr$. At search time, it is used during graph traversal but it is filtered out from the $k$ nearest neighbors result. Once a sufficient number of deletions is accumulated, they are performed in batches following the previously described steps. 

\begin{algorithm2e}[t]
\small
\DontPrintSemicolon
    \KwIn{graph $\Gr = (\V, \E)$, query $\q$, number of neighbors $k \in \Nat$, priority queue capacity $W \geq k$, initial candidates $\set{S} \subset \set{V}$, similarity function $\simfun$}
    \KwResult{$k$ approximate nearest neighbors to $\q$ in $\Gr$}
    
    $\queue = \set{S}$
    \tcp*[l]{Initialize candidate set $\queue$.}
    
    \tcp{Initially, no nodes are marked as explored.}
    \While{there exists an unexplored node in $\queue$}{
        $\vect{x} = $ closest unexplored node to $\q$ in $\queue$ w.r.t. $\simfun$\;
        Mark $\vect{x}$ as explored\;
        \lFor{$\vect{x}^{\prime} \in \set{N}(\vect{x})$}{
            $\queue \gets \queue \cup \vect{x}^{\prime}$
        }

        \tcp{Limit the size of $\queue$ to at most $W$:}
        $\queue = $ the (at most) $W$ closest nodes to $\q$ in $\queue$ w.r.t $\simfun$\;
    }
    \nllabel{alg:search_postprocessing}
    \Return{$k$ nearest nodes to $\q$ in $\queue$ w.r.t. $\simfun$}
    
    \caption{Greedy graph search.}
    \label{alg:search}
\end{algorithm2e}

\begin{algorithm2e}[t]
\small
\DontPrintSemicolon
    \KwIn{graph $\Gr$, $\vect{x} \in \V$, set $\Cl$ of out-neighbor candidates for $\vect{x}$, relaxation factor $\alpha \in \Real^+$, out-degree bound $\maxOutDeg \in \Nat$, similarity function $\simfun$}
    \KwResult{The new out-neighbors $\set{N}(\vect{x})$ of $\vect{x}$ in $\Gr$ s.~t.~$|\set{N}(\vect{x})| \leq \maxOutDeg{}$.}
    $\Cl \gets (\Cl \cup \set{N}(\vect{x})) \setminus \{ \vect{x} \}$
    \tcp*[l]{Add the current out-neighbors}
    
    $\set{N}(\vect{x}) \gets \emptyset$
    \tcp*[l]{Clear the out-neighbors of $\vect{x}$}
    
    \While{$\Cl \neq \emptyset$}{
        $\displaystyle \vect{x}^* \gets \argmax_{\vect{x}'' \in \Cl} \simfun(\vect{x}, \vect{x}'')$\;
        $\set{N}(\vect{x}) \gets \set{N}(\vect{x}) \cup \{ \vect{x}^* \}$\;
        \lIf{$| \set{N}(\vect{x})| = \maxOutDeg{}$}{
            \texttt{break}     
        }
        \For{$\vect{x}' \in \Cl$}{
            \lIf{$\alpha \cdot \simfun(\vect{x}^*,\vect{x}') \geq \simfun(\vect{x},\vect{x}')$}{
                \nllabel{alg:pruning_rule}
                $\Cl \gets \Cl \setminus \{\vect{x}'\}$
            }
        }
    }
    \caption{Neighborhood graph pruning~\cite{subramanya_diskann_2019}.}
    \label{alg:pruning}
\end{algorithm2e}

\section{Turbo LVQ}
\label{app:turbo_LVQ}

\begin{figure}
  \centering  
  \includegraphics[width=.95\textwidth]{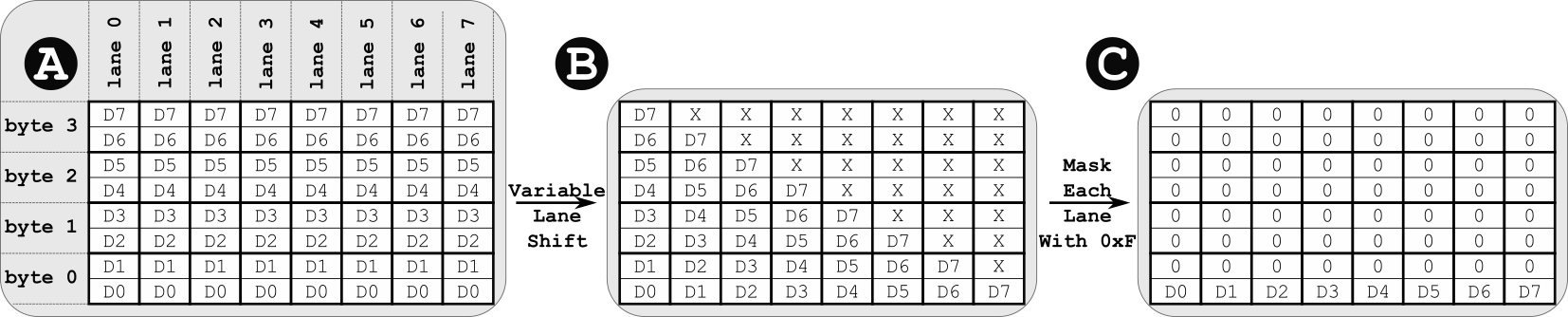}  
  \caption{The original LVQ implementation unpacks the representation in groups of eight dimensions by following a three-step procedure: (1) in block A, a 32-bit word containing eight 4-bit encodings is broadcasted into 8 lanes of a SIMD register;
    (2) a variable shift is applied to each lane in block B to get different dimensions into the four least significant bits of each lane;
    (3) finally, a lane-wise mask is applied (block C) to obtain each vector code as a 32-bit unsigned integer.}
  \label{fig:sequential_lvq_appendix}
\end{figure}  

\begin{figure}
  \centering  
  \includegraphics[width=0.6\textwidth]{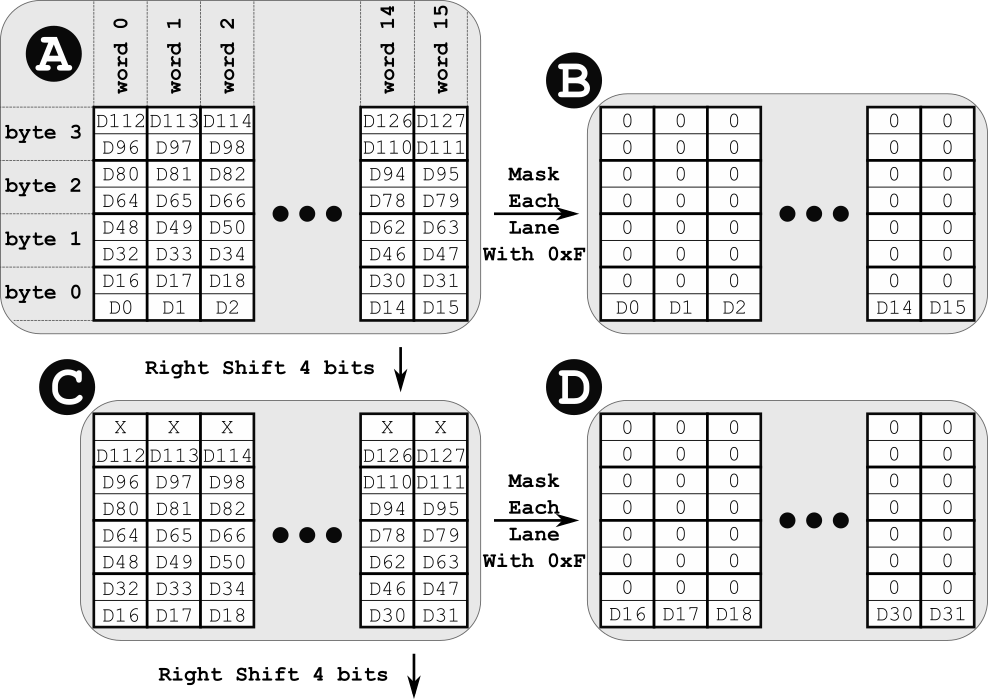}  
  \caption{Turbo LVQ uses SIMD registers in a more efficient way, by shuffling the dimensions so that they can be unpacked using a simplified workflow that only requires a right shift and a mask. }
  \label{fig:turbo_lvq_appendix}
\end{figure}  

In this work, we introduce a novel vector data layout for LVQ that achieves a 28\% search performance boost over the original version~\cite{aguerrebere_similarity_2023}, that we call hereafter Sequential LVQ. In this section, we describe the main ideas behind Turbo LVQ.

\subsection{Solution details}
\label{app:turbo_implementation}
The default LVQ implementation stores consecutive logical dimensions sequentially in memory.
While convenient, this choice requires significant effort to unpack.
To understand this, we explain the unpacking process (the process of extracting packed, encoded dimensions into a more useful form) for sequential LVQ.
This unpacked form is usually an 8 or 16-wide SIMD vector of integers.
Sequential LVQ uses an unpacking granularity of 8 shown in \cref{fig:sequential_lvq_appendix}.
First (block A), a 32-bit word containing eight 4-bit encodings is broadcasted into 8 lanes of a SIMD register.
A variable shift is applied to each lane (block B) to get different dimensions into the 4 least significant bits of each lane.
Finally, a lane-wise mask is applied (block C) to obtain each vector code as a 32-bit unsigned integer.
To obtain a 16-wide SIMD vector, this process needs to be applied to the next 8 dimensions and the two 8-wide registers are horizontally concatenated.
In total, every unpacking of 16 dimensions requires 7 assembly instructions.

With Turbo LVQ, we recognize that consecutive logical dimensions need not be stored consecutively in memory.
Instead, it can be better to store encoded dimensions permuted in memory to facilitate faster decompression with SIMD instructions.

\cref{fig:turbo_lvq_appendix} shows how we can accomplish this efficiently for 4-bit quantization by storing groups of 128 4-bit codes into 64-bytes of memory.
Block A shows the layout in memory.
Each 32-bit word contains 4 bytes and words are sequential in memory.
Each dimension encoding occupies 4-bits (i.e., a nibble).
Dimension 0 (D0) is stored in nibble 0 of word 0.
Dimension 1 (D1) is stored in nibble 0 of word 1 (beginning at an offset 4-bytes higher than D0).
This pattern continues until D16, which is stored in nibble 1 of word 0.
When decoding, the entire 64-block A is loaded into an AVX-512 register as 16 lanes of 32-bit integers.

Block B shows how we extract the first 16 dimensions (D0 to D15) from block A by simply applying a bitwise mask to each lane to retain the lowest 4 bits.
To get the next 16 dimensions, we apply a 4-bit right shift to each lane (block C) and again mask out all but the 4 least significant bits.
This process of shifting and masking continues until every group of 16 dimensions has been unpacked, at which point the next group of 128 dimensions is loaded into the register and the process repeats.
With this strategy, unpacking 16-dimensions requires only 2 assembly instructions: a load and mask for the first group and a shift-plus-mask for each following group, as opposed to the 7 instructions required for Sequential LVQ.

If a block 64-bit block is not completely filled with dimensions (which can occur when the number of dimensions is not a multiple of 128), we pad the remaining slots with zeros and terminate the shifting/masking procedure early and use masked SIMD operations for the final unpacked group.

The exact permutation to use depends on the number of lanes we wish to unpack at each iteration (16 in the above example) and the number of dimensions stored in each lane (8). For example, using 8-bit LVQ with AVX-512, we would still seek to unpack 16-lanes at a time, but would only have 4 dimensions per lane. If we were using AVX-512 extensions with native support for 16-bit floating point arithmetic, we might instead try to unpack 32 lanes at a time\cite{afroozeh_fastlanes_2023}.

\section{On the impact of LVQ in search performance}
\label{ssec:search_different_noise_levels}
To empirically establish the relationship between the reconstruction error level and the search window size required to achieve a target recall, we first compress the vectors using LVQ with different error levels (number of bits $B_1$ and $B_2$). Then, we use the first-level quantization to run the graph search for different search window sizes, and use the two-level vectors to re-rank the retrieved candidates, to finally compute the achieved search recall (see details in \cref{ssec:introduce_lvq}). 

In order to provide sub-bit granularity to the analysis, we generate the LVQ first-level ($Q(\vect{x})$) and two-level ($Q_{2L}(\vect{x}) = Q(\vect{x}) + Q_{res}(\vect{x})$) vector encodings using functions \texttt{LVQ\_firstLevel} and \texttt{LVQ\_twoLevel}, respectively (see code below). Note that this allows to use continuous number of bits $B_1$ and $B_2$, thus enabling sub-bit granularity. We save float32 versions of $Q(\vect{x})$ and $Q_{2L}(\vect{x})$ which we use to run the search with \svsfullp{}. First, the graph search is ran with $Q(\vect{x})$ $\forall \vect{x} \in \X$ to retrieve a set of neighbor candidates $\vect{x}^*_1,\dots,\vect{x}^*_K$. Next, the $Q_{2L}(\vect{x}^*_1), \dots, Q_{2L}(\vect{x}^*_K)$ are used to re-rank the $K$ retrieved candidates to finally provide the $k$ nearest neighbors and compute the achieved recall.  

\begin{lstlisting}[language=python]
import numpy as np
import numpy.typing as npt


def LVQ_firstLevel(Xin: npt.ArrayLike, B1: float) -> (npt.ArrayLike, float):

    # Center the input vectors
    X_means = np.mean(Xin, axis=0)
    X = Xin - X_means

    # Compute per-vector maximum and minimum across dimensions
    u = np.max(X, axis=1, keepdims=True)
    l = np.min(X, axis=1, keepdims=True)

    # Compute the quantization step and its inverse (avoiding division by zero)
    delta = (u - l) / (2 ** B1 - 1)
    delta[np.where(delta == 0)] = 1.0
    inv_delta = 1.0 / delta

    # Compute the codes
    codes = np.floor(np.multiply(inv_delta, X - l) + 0.5)
    codes = np.clip(codes, 0, (2 ** B1 - 1))

    # Compute the first-level reconstruction
    Xcomp = np.multiply(delta, codes) + l
    Xcomp += X_means
    return Xcomp.astype('float32'), delta


def LVQ_twoLevel(Xin: npt.ArrayLike, B1: float, B2: float) -> npt.ArrayLike:

    # Generate first-level compression
    Xcomp, delta = LVQ_firstLevel(Xin, B1)

    # Compute residuals
    X_res = Xin - Xcomp

    # Compute the quantization step and its inverse (avoiding division by zero)
    delta_res = np.array(delta, dtype='float32') / (2.0 ** B2)
    delta_res[np.where(delta_res == 0)] = 1.0
    inv_delta_res = 1.0 / delta_res

    # Compute the codes
    codes = np.round(np.multiply(X_res, inv_delta_res))
    codes = np.clip(codes, -2 ** (B2 - 1), 2 ** (B2 - 1) - 1)

    # Compute the two-level reconstruction
    Xcomp_res = np.multiply(codes, delta_res)
    Xcomp_res += Xcomp
    return Xcomp_res.astype('float32')
\end{lstlisting}

\section{Experimental Setup}
\label{sec:experimental_setup}

This section provides a detailed explanation of the experimental setup used in the evaluation presented in \cref{sec:experimental_evaluation}.

\subsection{Datasets}
\label{app:datasets}
In order to cover a wide variety of practical scenarios, we consider datasets of different scales ($n=10^6,\dots,10^8$) and dimensionalities ($d$=96, 512, 768) generated from diverse deep learning models. We evaluate on the standard dataset laion-img-512-1M, which corresponds to the first 1 million embeddings made available by the LAION-400M dataset~\cite{schuhmann_laion-400m_2021} generated from images using a multimodal model, on the recently introduced rqa-768-10M-ID generated from text snippets using a dense passage retriever~\cite{tepper_leanvec_2023} (we use both the 1M and 10M in-distribution versions of this dataset), and on classical similarity search datasets generated from images using a deep convolutional neural network (deep-96-1M and deep-96-100M correspond to the 1M and 100M first elements of the Deep1B dataset~\cite{babenko_deep_2016}). Moreover, we introduce a new dataset, open-images-512-13M, that allows us to benchmark our method under natural data distribution shift scenarios. The diversity of considered deep learning models (multimodal, text-only, image-only) provides solid empirical support to our contributions.

The datasets rqa-768-10M-ID, open-images-512-13M and Deep1B provide learning and query sets. We use the first 5000 vectors from the corresponding learning sets for parameter tuning. We use the provided query sets to report the final recall. For laion-img-512-1M, we use the first 10000 and following 5000 elements from the embeddings file number 101 of the LAION-400M dataset to build the query and learning sets, respectively.\footnote{The 400M embeddings in the LAION-400M dataset are organized in files containing chunks of 100M embeddings each.} 

\subsubsection{New dataset: open-images-512-13M}
\label{suppMat:open-images-dataset}
We introduce open-images-512-13M, a new dataset designed to recreate natural data distribution shifts. It consists of 13M CLIP~\cite{radford_learning_2021} embedding vectors ($d=512$) generated from images obtained from the Google’s Open Images~\cite{kuznetsova_open_2020} dataset. We take the 1.9M  images subset \footnote{\url{https://storage.googleapis.com/openimages/web/download_v7.html\#dense-labels-subset}} that includes dense annotations and use the provided bounding boxes to extract over 13M image crops with their corresponding class labels. There are a total of $434$ classes representing diverse objects. \cref{fig:open-images-dataset} shows examples of the image crops corresponding to four different classes (e.g., box, food, building and fountain), and the distribution of their vectors. We then generate CLIP embeddings for all the image crops and split them into a base vectors set with 13M elements, a query and a learning set with 10000 elements each. The open-images-512-1M corresponds to the first 1 million elements of the 13M base vectors. For both the query and learning sets with compute the corresponding 100 ground-truth nearest neighbors using maximum inner product. Although built for inner product, the vector embeddings are normalized, so the Euclidean distance can also be used. 

The code to generate the open-images-512-13M dataset can be found at \url{https://github.com/IntelLabs/VectorSearchDatasets}.

\subsection{Methods}
We compare \svs{} to the recently introduced streaming similarity search algorithm FreshVamana~\cite{singh_freshdiskann_2021} and the widely adopted HNSWlib~\cite{malkov_efficient_2020}. We use the Python packages publicly available at:

\begin{minipage}{\textwidth}
    \centering
    \begin{tabular}{ll}
        \toprule
        FreshVamana & \url{https://github.com/microsoft/DiskANN/blob/main/python/README.md} \\
        HNSWlib & \url{https://github.com/nmslib/hnswlib} \\
        \bottomrule
    \end{tabular}    
\end{minipage}

Inverted indices (IVF) combined with PQ are a widely adopted solution that becomes competitive in the high-throughput regime when implemented using GPUs~\cite{simhadri_results_2022,zhao_song_2020,ootomo_cagra_2023,johnson_billion-scale_2021}, or in CPU with distance computations optimized using AVX instructions (e.g., FAISS-IVFPQfs)~\cite{johnson_billion-scale_2021,guo_accelerating_2020}. The streaming techniques in FAISS, that are competitive in the high-throughput regime, do not reach an acceptable minimum accuracy (10 recall@10) of 0.9. They require re-ranking to achieve such high recall and the data structures used for re-ranking do not support deletions in the latest version (faiss-cpu 1.7.4) available at \url{https://pypi.org/project/faiss-cpu/} on \today.
Baranchuck et al.~\cite{baranchuk_dedrift_2023} recently introduced an approach to use inverted indices for streaming data with distribution shifts but their code is not publicly available (\today).

\subsection{Parameters setting}
For \svs{} and FreshVamana, we use the following parameter setting for graph building: $\maxOutDeg=64$ and $\maxOutDeg=128$ for datasets with 1M and over $10M$ vectors, respectively; $\alpha = 1.2$ and $\alpha = 0.95$ for Euclidean distance and inner product, respectively; and a search window size for building of 200. For HNSWlib, we use the same $\maxOutDeg$ setting as for the other methods\footnote{This corresponds to $M=32$ and $M=64$ in HNSW parameter notation.}, and a search window size for building of 500. LVQ-compressed vectors are always padded to half cache lines ($p=32$).

\subsection{Parameter tuning} \svs{} has three parameters that can be tuned to improve search performance: the search buffer capacity and two prefetching parameters (see \cref{sssec:graph_search}). We tune these parameters by doing an efficient grid search. To avoid overfiting, we use the vectors in the learning set as queries for parameter tuning and report the final recall vs. QPS results using the query set. 

The parameter tuning is done \textbf{only once} during the data streaming process, thus minimizing the overhead it introduces. In the case of the identically distributed data it is done at the first iteration. For the data distribution shift case, the tuning is done once the ramp-up finishes, as at that point the index achieved its maximum number of vectors. No tuning is done during ramp-up. In both cases, once tuned the parameters are kept fixed for subsequent iterations. 

Parameter tuning using the learning set requires computing the ground-truth by doing a linear scan with full precision vectors, which may be problematic in a real-world scenario. However, we empirically observed that this is not necessary in practice, as calibration is accurate enough when using an \emph{approximated} ground-truth, obtained by running the search in the graph with a large enough search window size ($1000$ for all our datasets). For the large scale datasets open-images-512-13M and rqa-768-10M-ID, where tuning makes an important difference, the total time taken by the \emph{approximated} ground-truth computation and parameter tuning is less than 10\% of the time taken by delete consolidations. For smaller scale datasets open-images-512-1M and rqa-768-1M-ID it is around 2x the delete consolidations time, which is still a minor cost considering it is done only once.

\subsection{System Setup}
\label{app:system_setup}
\begin{figure}
  \centering  
  \includegraphics[width=0.95\textwidth]{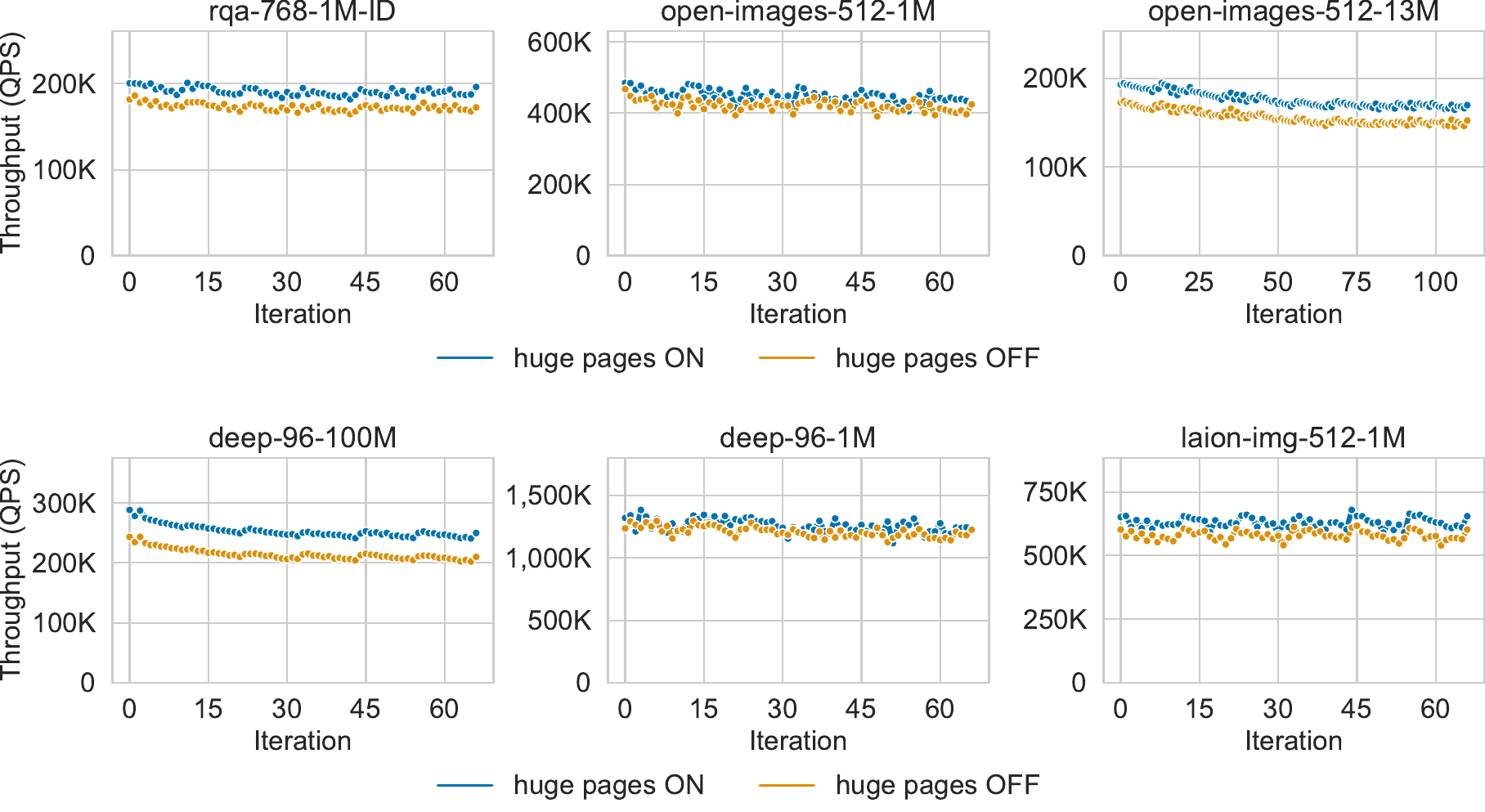} 
  \caption{Effect of the use of huge pages in throughput for IID streaming data (see the protocol in \cref{ssec:datasets_and_procols}) at iso-recall (0.9 10-recall@10) for \svs{}. \svs{} uses huge pages natively to reduce virtual memory overheads and results show a non-negligible performance boost when huge pages are enabled in the system.}
  \label{fig:bench_qps_over_time_huge_pages_svs_appendix}  
\end{figure}

\begin{figure}
  \centering  
  \includegraphics[width=0.95\textwidth]{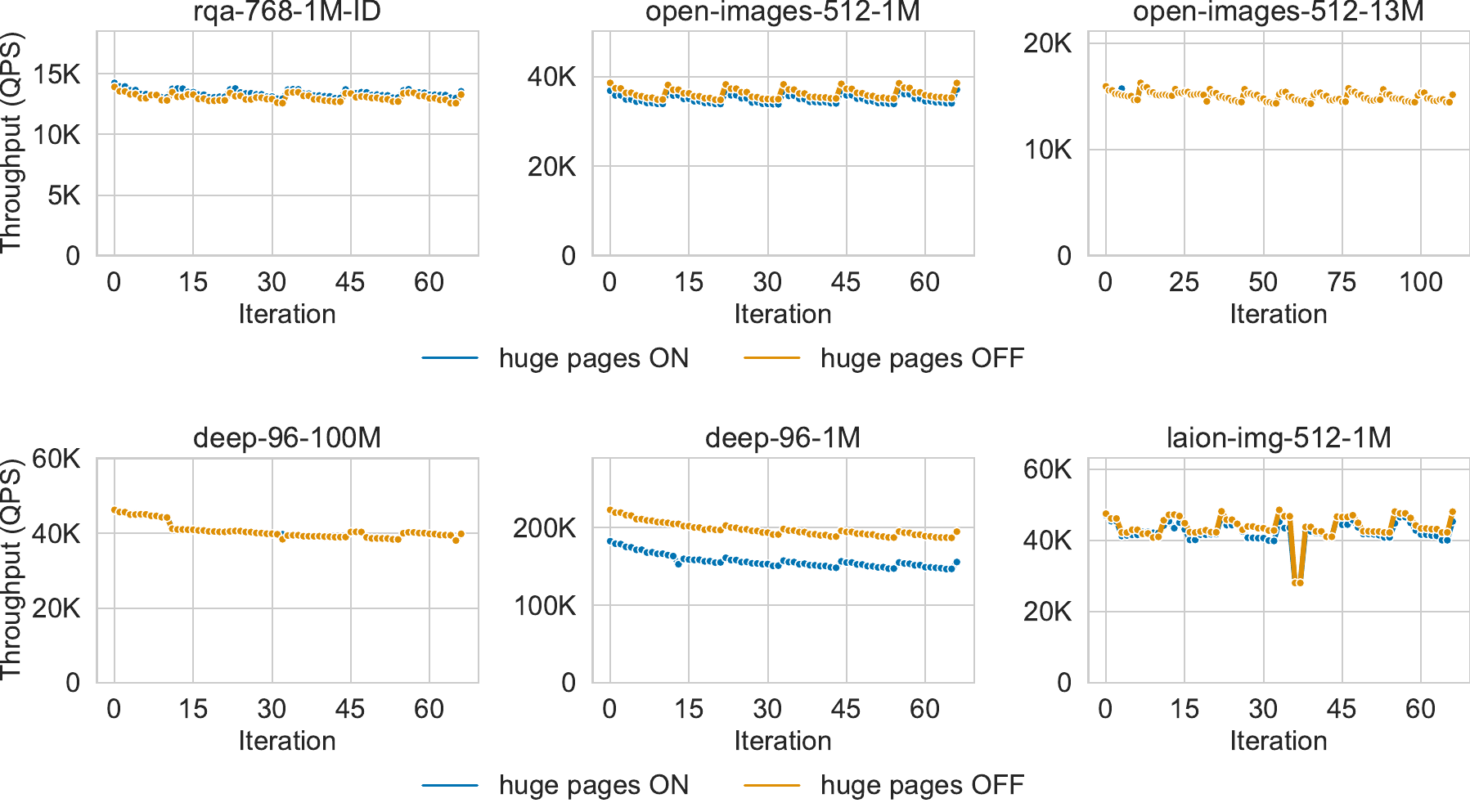} 
  \caption{Effect of the use of huge pages in throughput for IID streaming data (see the protocol in \cref{ssec:datasets_and_procols}) at iso-recall (0.9 10-recall@10) for FreshVamana. We ran FreshVamana with system flags enabled to automatically use huge pages for large allocations. Nevertheless, the method does not benefit from the use of huge pages for the evaluated datasets, and performance is degraded for the deep-96-1M dataset.}
  \label{fig:bench_qps_over_time_huge_pages_freshVamana_appendix}  
\end{figure}

\begin{figure}
  \centering  
  \includegraphics[width=0.95\textwidth]{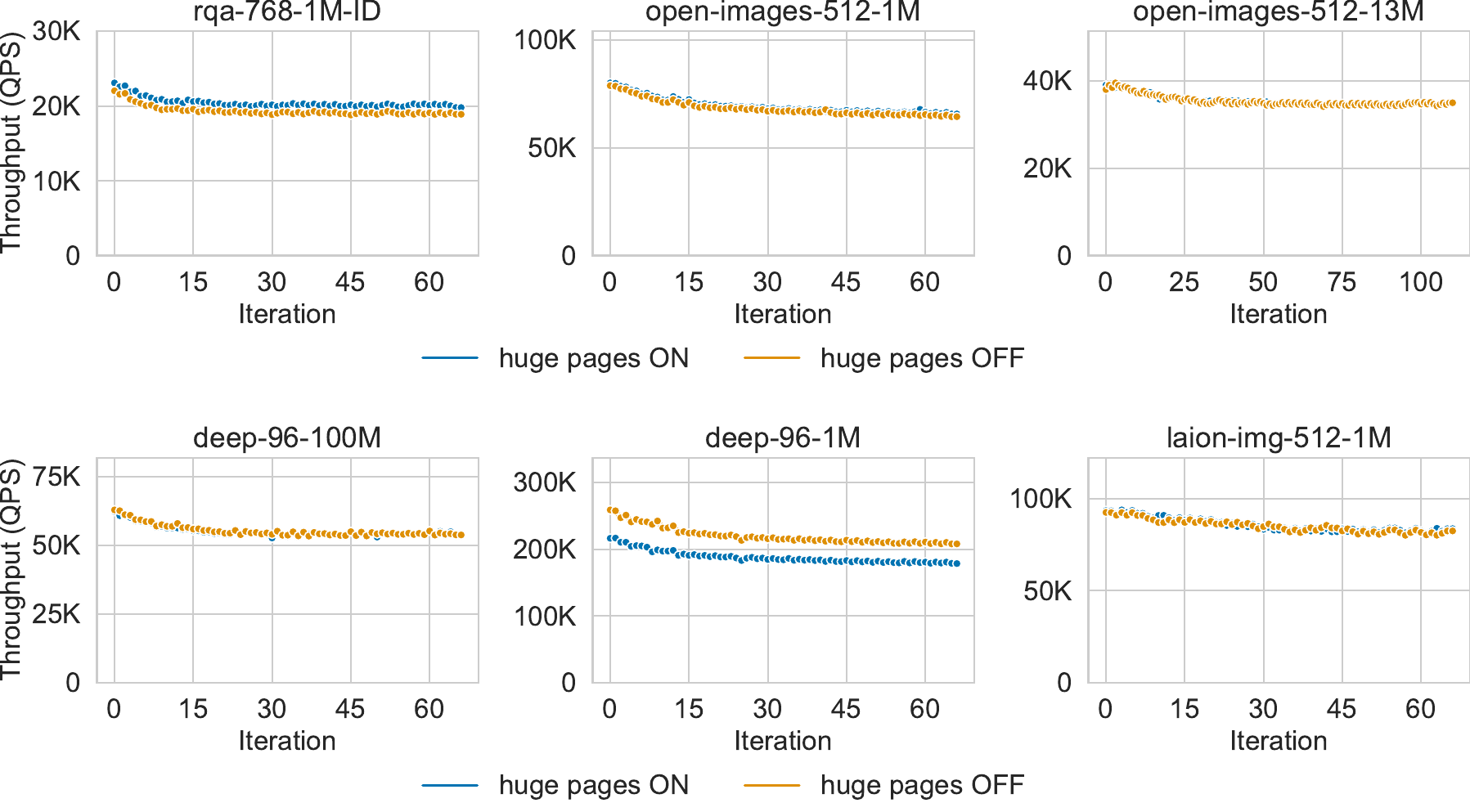} 
  \caption{Effect of the use of huge pages in throughput for IID streaming data (see the protocol in \cref{ssec:datasets_and_procols}) at iso-recall (0.9 10-recall@10) for HNSWlib. We ran HNSWlib with system flags enabled to automatically use huge pages for large allocations. Nevertheless, the method does not benefit from the use of huge pages for most of the evaluated datasets, and performance is degraded for the deep-96-1M dataset.}
  \label{fig:bench_qps_over_time_huge_pages_HNSW_appendix}  
\end{figure}

We use the \textit{hugeadm} Linux utility to preallocate 1GB huge pages. \svs{} uses huge pages natively to reduce virtual memory overheads. We run the other methods with system flags enabled to automatically use huge pages for large allocations for a fair comparison. In some cases, we noticed that enabling huge pages is detrimental to FreshVamana's and HNSWlib's performance. Figures \ref{fig:bench_qps_over_time_huge_pages_svs_appendix} to \ref{fig:bench_qps_over_time_huge_pages_HNSW_appendix} show a comparison of the throughput achieved with and without huge pages by \svs{}, FreshVamana and HNSWlib, respectively, for various datasets.

\svsfullp{} supports the option of setting the dimensionality at compile time (static) versus at runtime (dynamic), which provides a search performance boost~\cite{aguerrebere_similarity_2023}. We enabled static dimensionality for the dimensionalities of the considered datasets. See \url{https://intellabs.github.io/ScalableVectorSearch/advanced/static_dimensionality.html} for details on how to enable static dimensionality for SVS. 

\subsection{M-LVQ implementation}
\label{app:mlvq_implementation}
A script showing how to run M-LVQ using the SVS library can be found at \url{https://github.com/IntelLabs/ScalableVectorSearch/tree/main/reproducibility}.

\section{Complementary Experimental Results}
\label{app:extra_results}
\begin{figure}
  \centering  
  \includegraphics[width=0.9\columnwidth]{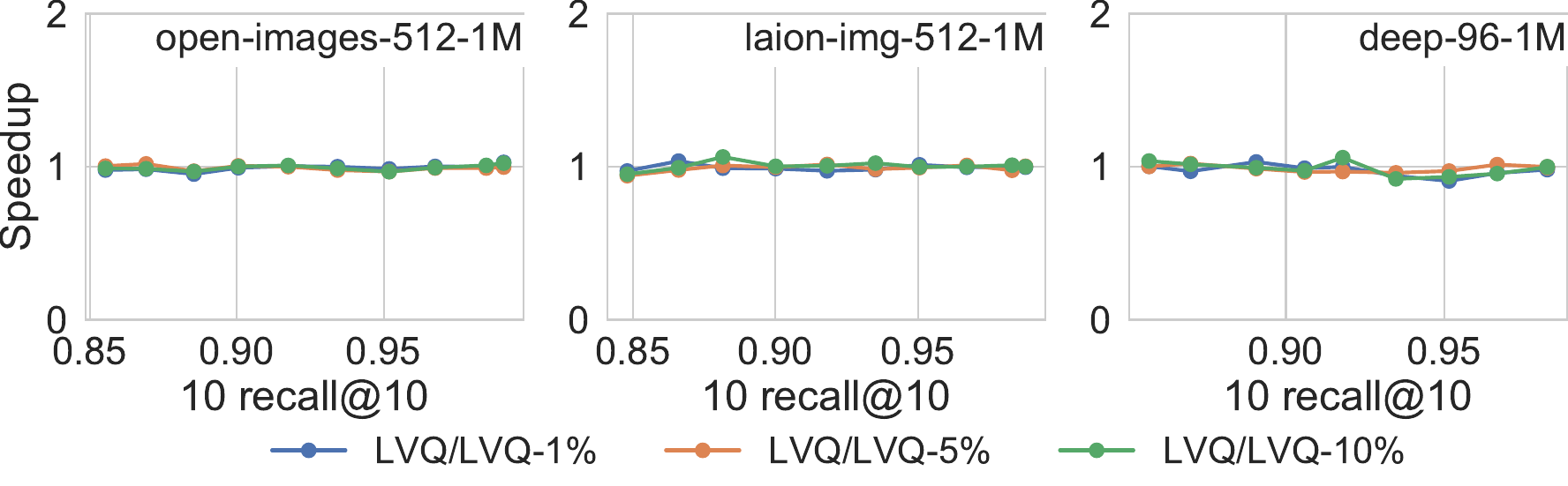}  
  \caption{Similarity search with \svs{} is robust to noisy estimates of the mean. For three datasets and different recall values, we plot the speedup obtained by computing $\vect{\mu}$ from the entire dataset with respect to the throughput obtained when using random samples of 1\%, 5\% and 10\% to compute $\vect{\mu}$. The ratio is 1 for all cases, showcasing LVQ's robustness: computing the mean with as few as 1\% of the vectors is enough to achieve peak performance. The search window size required to achieve each target recall remains unchanged when using as few as 1\% of the vectors.}
  \label{fig:LVQ_robustness_appendix}
\end{figure}

\begin{figure}
  \centering  
  \includegraphics[width=0.95\textwidth]{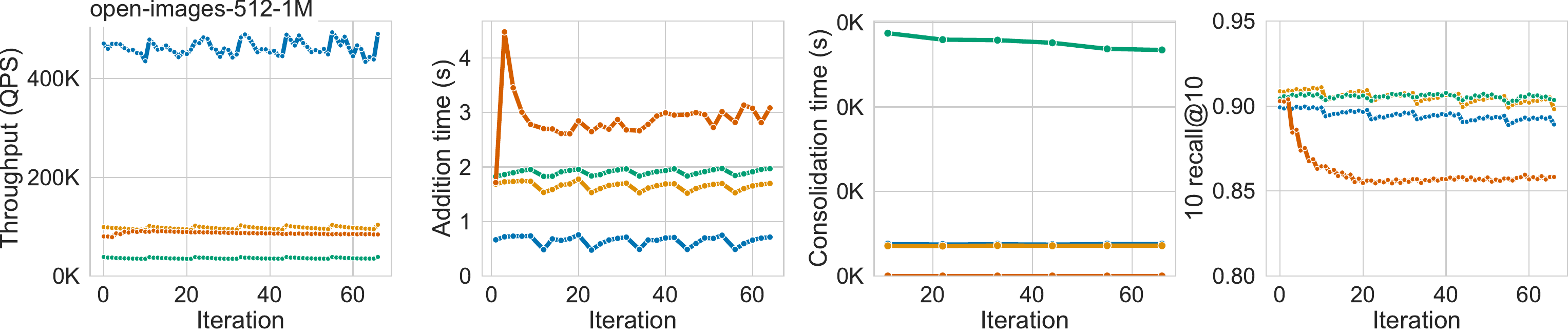} 
  \includegraphics[width=0.95\textwidth]{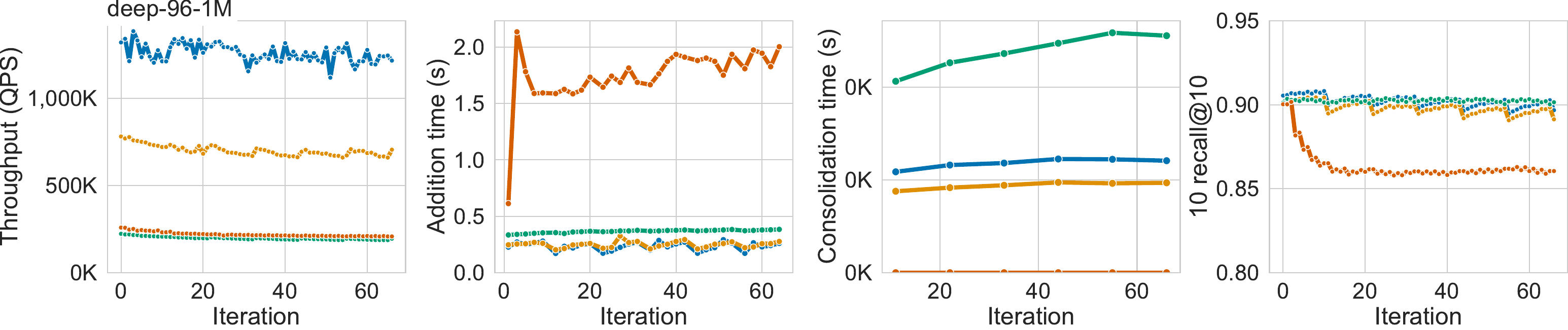} 
  \includegraphics[width=0.95\textwidth]{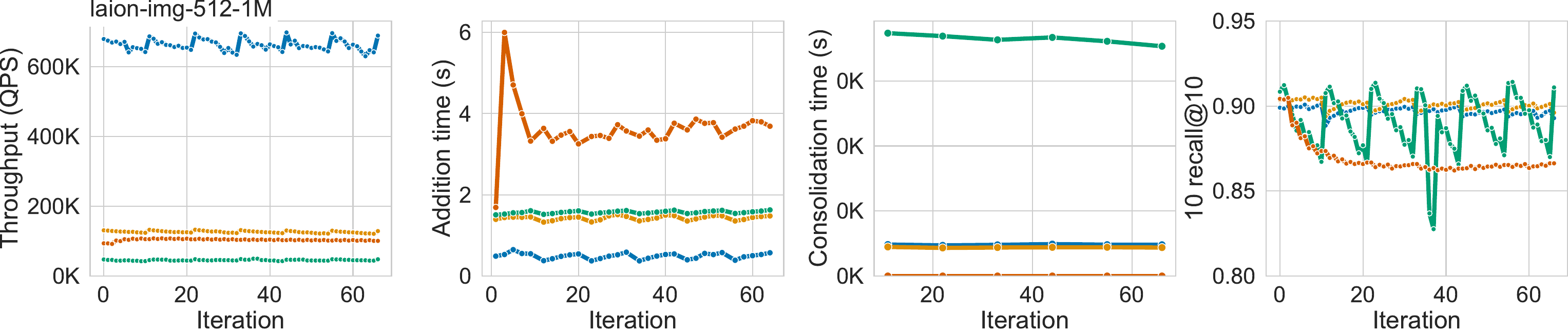} 
 \includegraphics[width=0.95\textwidth]{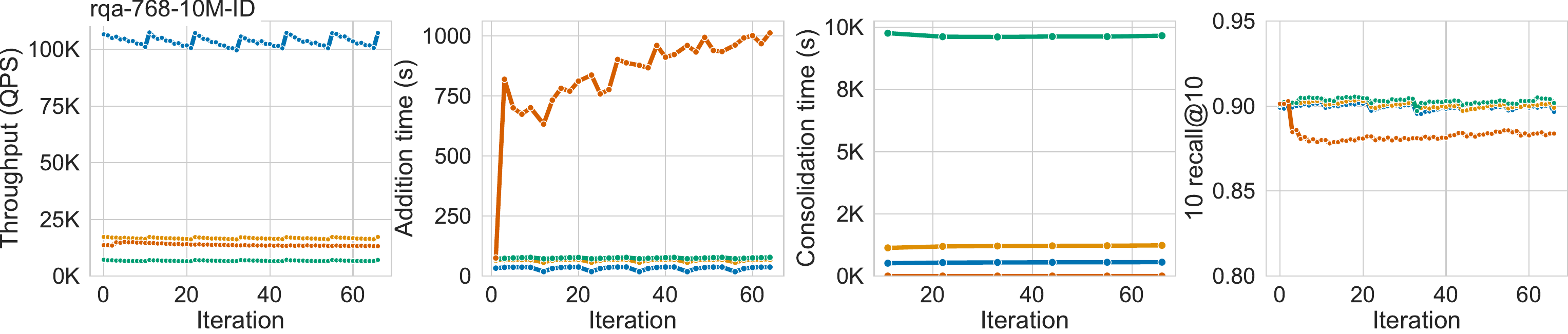} 
    \includegraphics[width=0.65\textwidth]{figs/bench_iid_over_time_legend.pdf} 
  \caption{Performance comparisons for IID streaming data (see the protocol in \cref{ssec:datasets_and_procols}). From left to right: search throughput at iso-recall (10-recall@10 of 0.9), runtime for the batched additions, runtime for the delete consolidation rounds, and recall for a fixed search window size (set to achieve 0.9 10-recall@10 at $t=0$). HNSWlib does not perform delete consolidations (the reported value is zero), causing a degradation in recall over time (rightmost column). \svsfullp{}{} manages to maintain a stable recall over time. We observe an unexpected behavior for the laion-img-512-1M dataset when using the FreshVamana implementation. The recall decreases with the vector additions and deletions between consolidations, exhibiting a sawtooth pattern. }
  \label{fig:bench_qps_over_time_appendix}  
\end{figure}

\begin{figure}
  \centering  
  \includegraphics[width=0.5\columnwidth]{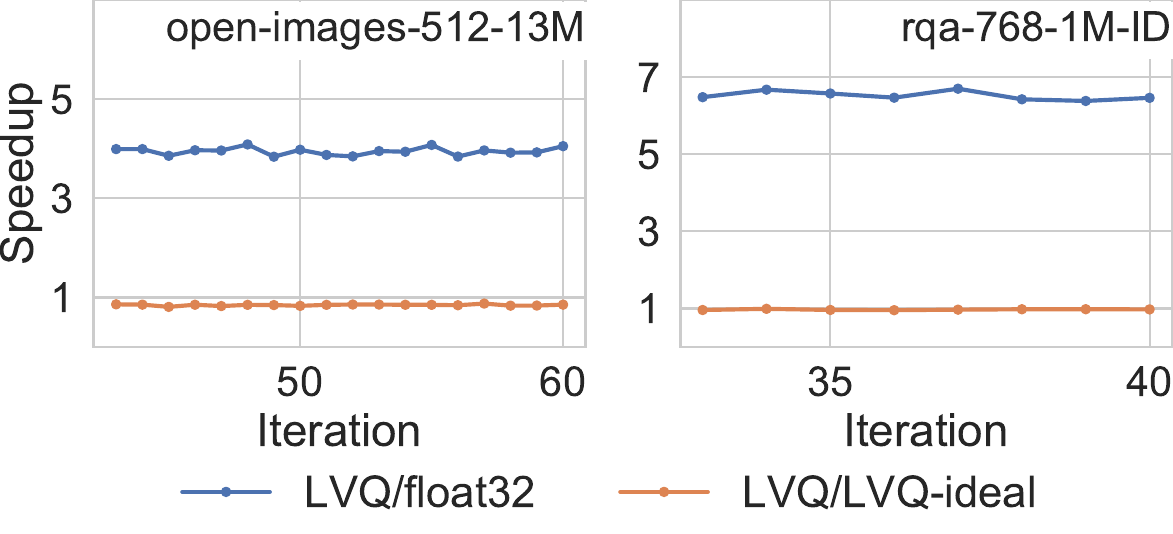}
  \caption{Search performance speedup for streaming data during the steady-state (see the protocol in \cref{ssec:datasets_and_procols}). \svs{} achieves significant speedups over SVS-float32 (blue curve) under data distribution shifts (see the protocol in \cref{ssec:datasets_and_procols}). When compared to SVS-LVQ-ideal, which represents the ideal and impractical scenario where the entire database is re-encoded at each $t$ using LVQ with the sample mean from $\X_t$, the constant speedup of 1 in the orange curve indicates no degradation in search performance.}
  \label{fig:LVQ_robustness_distro_shift_appendix}
\end{figure}

\begin{figure}
  \centering    
  \includegraphics[width=0.85\columnwidth]{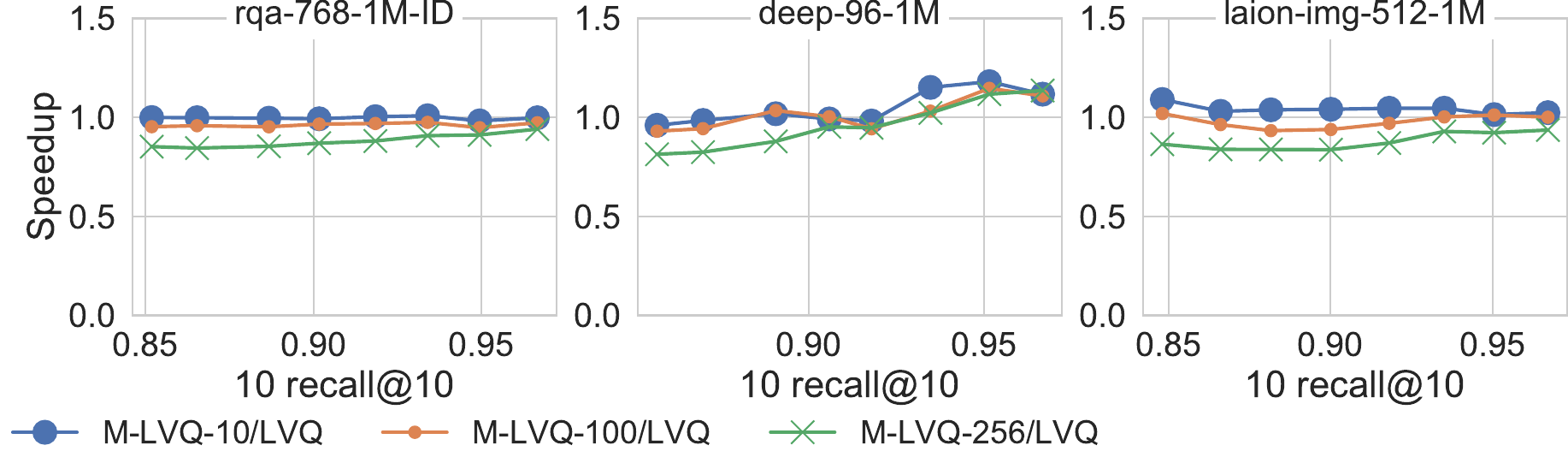}    
  \caption{Search performance speedup achieved with M-LVQ over LVQ using $M=10,100,256$ for static indexing ($B_1=4$, $B_2=8$). The rqa-768-1M-ID dataset does not benefit from using multiple means. The deep-96-1M dataset benefits for the very high recall regime (>0.94). The laion-img-512-1M dataset benefits slightly with a small number of means ($M=10$).
  }
  \label{fig:LVQ_M-LVQ_appendix}
\end{figure}

\begin{figure}
  \centering  
  \includegraphics[width=0.5\columnwidth]{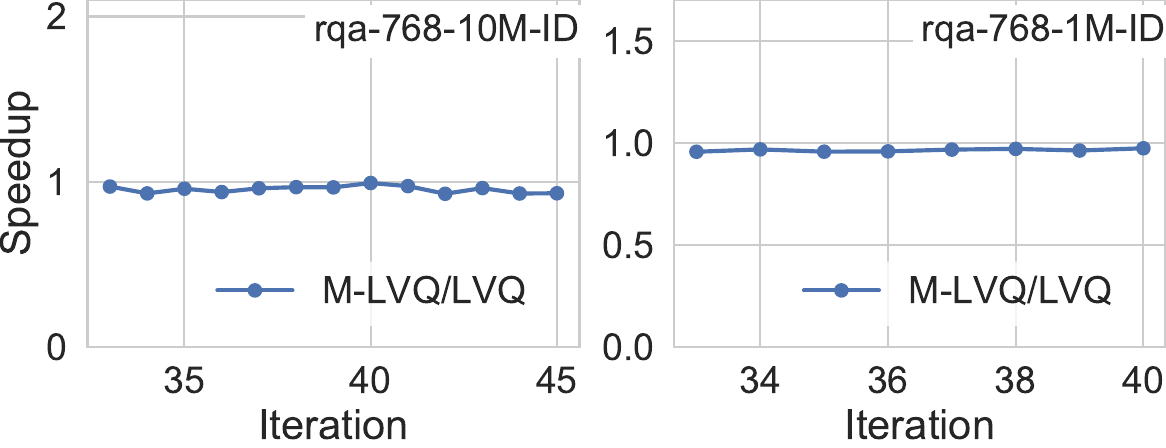} 
  \caption{Search performance speedup obtained by M-LVQ over LVQ using $M=100$ at 0.9 10 recall@10 accuracy during the steady-state phase, where additions and deletions are performed while keeping the total number of vectors fixed (see the protocol in \cref{ssec:datasets_and_procols}). No speedup, nor performance degradation, is observed for rqa-768-1M-ID and rqa-768-10M-ID.}
  \label{fig:LVQ_MLVQ_dynamic_appendix}
\end{figure}

Figures \ref{fig:LVQ_robustness_appendix} to \ref{fig:LVQ_MLVQ_dynamic_appendix} in this section show more experimental results supporting the experimental evaluation in \cref{sec:experimental_evaluation}.

\end{document}